\documentclass[preprint,12pt,authoryear]{elsarticle}
\usepackage{amsmath,amssymb,bm}
\usepackage{graphicx}
\usepackage{booktabs}
\usepackage[hidelinks]{hyperref}
\newtheorem{proposition}{Proposition}
\journal{ISPRS Journal of Photogrammetry and Remote Sensing}

\begin{document}
\begin{frontmatter}

\title{Spaceborne differential photogrammetry for control-free measurement of large-gradient deformation with structural immunity and a predictable accuracy envelope}
\tnotetext[tnote1]{Submitted to \textit{ISPRS Journal of Photogrammetry and Remote Sensing}.}

\author[szu]{Yueqiang Zhang}
\author[szu]{Chang Ma}
\author[szu]{Shuixin Pan\corref{cor1}}
\ead{shuixinpan@szu.edu.cn}
\author[hnu]{Haibo Liu}
\cortext[cor1]{Corresponding author.}
\affiliation[szu]{organization={State Key Laboratory of Radio Frequency Heterogeneous Integration; Key Laboratory of Optoelectronic Devices and Systems of Ministry of Education and Guangdong Province; Shenzhen Key Laboratory of Intelligent Optical Measurement and Detection; College of Physics and Optoelectronic Engineering, Shenzhen University}, city={Shenzhen}, postcode={518060}, country={China}}
\affiliation[hnu]{organization={School of Artificial Intelligence and Robotics, Hunan University}, city={Changsha},
postcode={410082}, country={China}}

\begin{abstract}
Optical satellite image correlation measures wide-area deformation in regimes where coherent interferometric synthetic aperture radar fails because displacement gradients are too large. However, standard pairwise workflows lack a pre-acquisition error budget and rely on extensive stable terrain. We formulate repeat-pass optical correlation as a differential estimation problem without surveyed ground control. Nominal georeferencing defines the coordinate frame, stable-area constraints and displacement priors resolve the datum, and surface displacement is estimated jointly with inter-epoch revisit-bias coefficients. The model yields a predictive accuracy envelope and calibrated per-point posterior uncertainty, bounds along-track uncertainty through a displacement prior, and represents pushbroom jitter using per-line revisit offsets. Simulations and Sentinel-2 and WorldView-2 experiments on the 2019 Ridgecrest earthquake, the 2023 Kahramanmara{\c{s}} earthquake, and the Baltoro glacier validate the predicted noise floor, control-free accuracy margin, and leakage caused by view-angle and digital elevation model errors. The measured noise floor reaches approximately $0.05$ pixel at $10$\,m ground sampling distance. With only five stable tiles, conventional destriping changes the estimated Baltoro trunk velocity from $106$ to $1251$\,m\,yr$^{-1}$, whereas the prior-constrained estimate remains $87$\,m\,yr$^{-1}$. Closure analysis attributes approximately $88\%$ of pair-error variance to individual scenes, consistent with $25{,}354$ ITS\_LIVE glacier-velocity triplets. Three matching methods lead to the same conclusions. The framework therefore turns pairwise correlation into a robust measurement with a predictive error budget, reduced dependence on stable terrain, and conclusions independent of the matching method.
\end{abstract}

\begin{keyword}
optical image correlation \sep differential photogrammetry \sep
surface deformation \sep repeat-pass satellite imaging \sep
error budget \sep uncertainty quantification
\end{keyword}

\end{frontmatter}

\section{Introduction}
\label{sec:intro}

Measuring the displacement of a surface from two images taken at different times is one of the oldest tasks of photogrammetry. 
High-precision digital image correlation was established in the photogrammetric literature four decades ago \citep{ackermann1984}.
Its subsequent application to satellite imagery enabled measurements of earthquakes from SPOT scenes \citep{vanpuymbroeck2000,michel2002} and measurements of glacier and permafrost flow from repeat optical acquisitions \citep{kaab2000,berthier2005}. 
These developments established optical correlation as a geodetic imaging technique \citep{avouac2014}.
Photogrammetry supplied sub-pixel matching methods together with an understanding of how their precision depends on image texture and window size \citep{debellagilo2011}. 
Optical deformation products, however, still lack a complete observation model in which inter-epoch change is the unknown, revisit geometry is estimated explicitly, and accuracy is predicted before acquisition and verified afterwards. Such treatment has long been standard in photogrammetric block adjustment \citep{fraser1997,luhmann2016}.

The scientific demand for such measurements is substantial. 
Surface deformation is one of the few geophysical observables that can be measured repeatedly from orbit over entire regions. 
It is central to studies of earthquakes, glacier dynamics, and landslide hazards.
Coseismic slip distributions recovered from space constrain rupture physics and update seismic hazard within days of an event \citep{barnhart2019,barbot2023}. 
Glacier velocity fields provide the dynamic component of the ice-mass budget. 
Glaciers in High Mountain Asia have slowed by approximately one tenth over two decades as they have thinned \citep{dehecq2019}, while global glacier mass loss has accelerated \citep{hugonnet2021}. 
The first global inventory of glacier velocity and thickness relies on repeat optical and radar imagery \citep{millan2022}. 
Slow-moving landslides, whose acceleration precedes catastrophic failure, are now tracked from space as an early warning quantity \citep{lacroix2019}. 
In each case the measurand is a \emph{change} between two acquisitions, the areas are large, and no operator can lay ground control across a fault zone, an accumulation basin, or a disaster site.

Two techniques serve this need and are complementary by construction.
Interferometric synthetic aperture radar (InSAR) can achieve millimeter-per-year precision over coherent terrain undergoing slow deformation \citep{insar_review}. 
Its performance degrades when displacement is large or spatial gradients are steep because dense fringes complicate phase unwrapping and surface changes reduce coherence \citep{liu2026coal,chen2026edge}. 
These limitations affect measurements of meter-scale coseismic offsets, rapidly moving landslides, and glacier flow of hundreds of meters per year. 
Optical image correlation instead estimates sub-pixel displacement directly from a repeat image pair \citep{avouac2014}. 
It has no phase ambiguity and provides two horizontal components rather than one line-of-sight component.
Since COSI-Corr \citep{leprince2007} it has become the workhorse of coseismic slip mapping \citep{milliner2015,milliner2020,han2023menyuan}, of landslide kinematics \citep{chang2024stacking}, and of glacier velocity at continental scale through GoLIVE \citep{fahnestock2016}, autoRIFT/ITS\_LIVE \citep{lei2021}, and Sentinel-2 processing chains \citep{kaab2016,altena2019}.

Yet, a decade and a half after \citet{leprince2007}, optical practice remains a \emph{procedure} rather than a measurement model. 
The standard pipeline correlates the pair, then removes an orbital ramp and along-track striping by fitting low-order surfaces and per-column medians on stable ground picked by hand \citep{scherler2008,stumpf2018,zhou2025corner}; the accuracy actually attained is established afterwards, empirically, by comparing against GNSS, field offsets, or an independent velocity product \citep{heid2012,zheng2023glaft}. 
Three consequences follow. 
The pipeline generally reports a displacement field without an error budget that predicts accuracy before image acquisition or supports self-consistency assessment afterwards. 
Its uncertainty output is usually limited to a proxy derived from the correlation peak, although downstream slip inversion, ice-flux estimation, and landslide warning require calibrated uncertainties. 
Existing workflows also lack a unified interface for incorporating sparse control, a digital elevation model (DEM), precise orbit determination, and redundancy across epochs or spectral bands. 
These sources of information are commonly introduced through separate post-processing steps.
For the science these products feed, the cost is concrete: velocity mosaics are stacked over years to beat down errors that a model would budget, and the seasonal and interannual signals of interest \citep{dehecq2019} sit inside the very scatter the stacking removes.

We argue that these are not incidental gaps but symptoms of a missing model, and that the model is available. 
Our preceding studies developed a differential formulation for vision-based deformation measurement. 
Instead of reconstructing the geometry of each epoch independently and then differencing the products, the formulation estimates inter-epoch change directly. 
This allows shared systematic errors to be canceled or absorbed during estimation. 
The first study introduced a differential six-degree-of-freedom pose estimator with first-order immunity to camera-calibration errors \citep{pamidraft}. 
The second recovered arcsecond-level relative motion under incomplete or absent control fields \citep{lian2026}.
The present work extends that framework to repeat-pass satellite imaging and shows that it converts optical deformation measurement from a procedure into a predictable measurement whose accuracy is known before acquisition and whose matching uncertainty is calibrated per pair after it.

A repeat-pass satellite acquisition has the structure of a revisit measurement. 
Nominal georeferencing plays the role of the pose prior that real-time kinematic positioning provides for a returning uncrewed aerial vehicle. The inter-orbit attitude bias acts as a revisit offset, while ground control is often sparse or absent.
From this perspective, rational polynomial coefficient (RPC) bias compensation, sub-pixel correlation, and destriping can be interpreted as components of a differential parameterization. 
Supplying the theory yields the error budget, the per-point posterior uncertainty, the fusion interface, and the honest information boundary that the procedure cannot.

The main contributions of this study are as follows.
\begin{enumerate}
\item We formulate a differential measurement model for repeat-pass optical imaging (Section~\ref{sec:method}). 
The model absorbs inter-orbit biases, bounds uncertainty along the imaging track through a displacement prior, and introduces a per-line revisit offset $\xi(t)$ for pushbroom attitude jitter.
\item We derive a closed-form accuracy envelope (Section~\ref{sec:sim}) for the matching-noise floor, the accuracy margin under control-free operation, the sensitivity to displacement-prior mismatch, and the leakage caused by view-angle and DEM errors. Monte Carlo simulations verify each component.
\item We validate the model without ground control using three real-data cases (Section~\ref{sec:real}). 
They comprise Sentinel-2 imagery of the 2019 Ridgecrest earthquake, WorldView-2 imagery of the
2023 Kahramanmara{\c{s}} earthquake, and Sentinel-2 imagery of the Baltoro glacier. 
The experiments test two resolution classes and two deformation regimes. 
They verify the predicted noise floor, closure-based error decomposition, pair-specific matching-uncertainty calibration, effects
of cross-view acquisition, and agreement with ITS\_LIVE velocities.
\item We show that conventional destriping is a robust special case of the differential estimator (Section~\ref{sec:duel}). 
Controlled thinning experiments and natural image cut-outs identify its failure when stable terrain is absent or moving terrain dominates the scene. The prior-constrained estimator remains operational in these cases. 
We also identify the along-track mean of a dense deformation field as an unobservable quantity for any control-free optical method.
\end{enumerate}

\section{Related Work}
\label{sec:related}

\subsection{Optical offset tracking: from COSI-Corr to continental products}
The photogrammetric lineage of the matcher runs from digital image
correlation \citep{ackermann1984} through the first satellite
applications \citep{vanpuymbroeck2000,michel2002,kaab2000,berthier2005}
to systematic precision studies of normalized cross-correlation (NCC) on
mass movements \citep{debellagilo2011}, aerial-photograph
co-registration for deformation \citep{ayoub2009}, and Pl\'eiades
stereo displacement of landslides \citep{stumpf2014}. Sub-pixel image correlation for ground displacement was formalized by \citet{leprince2007}, whose COSI-Corr workflow combining orthorectification, coregistration, and correlation established a processing template that remains widely used;
\citet{avouac2014} review its geodetic applications. Alpine-terrain
accuracy and its improvement by orthorectification and stable-ground
correction were quantified by \citet{scherler2008}, and the matching
algorithms themselves were benchmarked at global scale by
\citet{heid2012}. Open-source implementations followed
\citep[MicMac;][]{rosu2015}. Full-archive glacier velocities with a formal uncertainty treatment
were derived by \citet{dehecq2015} for the Pamir--Karakoram--Himalaya,
and multi-sensor regional mapping by \citet{millan2019}. Cross-correlation
feature tracking has also been used to map glacier velocity and identify
surge behavior in the Karakoram \citep{zhang2024karakoram}.
Glacier-velocity processing subsequently developed into operational
large-scale products. GoLIVE mapped ice flow over entire continents
using Landsat~8 \citep{fahnestock2016}, autoRIFT/ITS\_LIVE made the processing
autonomous and cloud-scale \citep{lei2021}, and Sentinel-2's five-day
revisit enabled weekly flow estimates \citep{kaab2016,altena2019};
GLAFT provides a community test kit for validating such products
\citep{zheng2023glaft}. Recent methodological work pushes temporal
stacking for slow signals \citep{chang2024stacking}, deep-learning
sub-pixel estimators \citep{montagnon2024deep}, corner-based
robustification \citep{zhou2025corner}, and the systematic study of how
image and DEM resolution limit accuracy \citep{antoine2025topo}. 
These methods generally operate on individual image pairs and apply
empirical corrections based on trend removal, destriping, and stable-
ground calibration. Accuracy is then assessed through external
comparison. 
What none provides is an estimation-theoretic account of the correction step, a pre-acquisition error budget, or a per-point posterior uncertainty with a calibrated matching component; that is the gap this paper closes, and it does so \emph{without} changing the matching kernel, so the result is a drop-in model for the existing chains.

\subsection{Coseismic and landslide applications}
Optical correlation supplied the near-field slip and off-fault
deformation of Landers \citep{milliner2015}, and for Ridgecrest 2019
daily PlanetScope imagery separated the two mainshocks
\citep{milliner2020} while joint geodetic inversions constrained slip
and stressing \citep{barnhart2019}---the event and field data we use for
validation. For the 2023 Kahramanmara\c{s} doublet, slip distributions
were derived from combined optical, InSAR and GNSS data
\citep{barbot2023}. Slow-moving landslides are tracked by the same
technique \citep{lacroix2019,chang2024stacking}, including precursory
motion before failure from Sentinel-2 \citep{lacroix2018} and
time-series inversion of optical displacement fields
\citep{bontemps2018}; ICA-based removal of image-geometry artefacts
\citep{aati2022} and the documented Landsat-8/Sentinel-2
misregistration \citep{storey2016} show that the geometry, not the
correlator, is the residual error source---the premise of this
paper; and Sentinel-2/Landsat
co-registration for surface-motion measurement has been refined
specifically for that purpose \citep{stumpf2018}. Across these
applications the two quantities repeatedly requested by the downstream
inversion---an error covariance and a stability check across
epochs---are exactly what a differential model delivers.

\subsection{InSAR and its large-gradient boundary}
Time-series InSAR (PS/SBAS) is the precision standard for slow, coherent deformation \citep{insar_review}. 
Its failure at high fringe gradient is an active research front---phase-unwrapping networks and model-driven unwrapping \citep{chen2026edge,rouetleduc2025unwrap}, and explicit fusion with pixel-offset tracking to bridge the large-gradient core \citep{liu2026coal}---which is precisely an admission that the large-gradient regime belongs to offset methods.
Our contribution addresses a different problem. We formulate a measurement model for the optical offset regime and provide per-point posterior uncertainties that can be used as weights in joint optical and InSAR inversion.

Direct validation of the near-field optical result using InSAR is not feasible for this event because the two techniques have different sensitivity ranges.
On the two LiCSAR tracks spanning the Ridgecrest sequence, coherence along the
mapped rupture trace has a median of $0.067$ (ascending 064A) and
$0.055$ (descending 071D), against scene medians of $0.659$ and
$0.827$: the interferogram is decorrelated exactly where the
displacement is large, so the unwrapped phase there carries no
information to compare against. Retreating to the mid-field does not
help either, because there the coseismic displacement is a few
centimetres, an order of magnitude below the optical noise floor
reported in Section~\ref{sec:real}. 
The two techniques therefore have largely disjoint sensitivity ranges for this event class. 
InSAR retains coherence mainly where motion is below the optical offset-tracking noise floor, whereas optical
correlation provides a measurable signal near the rupture where InSAR has decorrelated. 
The techniques are therefore complementary rather than suitable for direct mutual validation in the near field.
Validation of the near-field result is therefore carried by the independent field survey of surface offsets (Section~\ref{sec:ridgecrest}), not by InSAR.

\subsection{Pushbroom geometry and jitter}
Attitude jitter of pushbroom platforms is a known geometric-accuracy
limiter, addressed by attitude-data reimaging and multi-source
compensation \citep{wang2025haiyang,xia2024jitter,li2025dynamic}, and
Sentinel-2's own band-to-band and along-track geometric residuals have
been characterized as a limit on motion measurement
\citep{kaab2016,stumpf2018}. These are forward-model corrections of a
single acquisition; we show (Section~\ref{sec:pushbroom}) that in the
differential setting the per-line jitter is absorbed by a time-varying
revisit offset whose identifiability comes from the displacement
prior---an estimation view that complements the forward-modeling one.

\subsection{Differential vision metrology}
Close-range photogrammetry treats calibration and orientation as
parameters of an adjustment with a stated precision
\citep{fraser1997,luhmann2016}; the differential formulation used
here carries that discipline to the change measurement itself. This paper is the spaceborne instance of the differential
formulation introduced for fixed cameras \citep{pamidraft,lian2026}.
The screened-saturation effect of a displacement prior on an
along-track chain is derived here (Section~\ref{sec:screen}) for the
satellite geometry and the pushbroom line rate.

\section{Differential Model for Repeat-Pass Imaging}
\label{sec:method}

\subsection{Observation model}
For $N$ ground points observed in two repeat-pass images, stacking the
two ground components of every sub-pixel match into
$\Delta\bm u\in\mathbb R^{2N}$ gives
\begin{equation}
\Delta\bm u=\bm d+\bm B\bm\xi+\bm G\bm e+\bm n,
\label{eq:obs}
\end{equation}
with the displacement field $\bm d\in\mathbb R^{2N}$, revisit-bias
coefficients $\bm\xi\in\mathbb R^{p}$, DEM height errors
$\bm e\in\mathbb R^{N}$, and matching noise
$\bm n\sim\mathcal N(\bm 0,\bm\Sigma_n)$. Its per-point standard
deviations are $\sigma_{n,j}=\sigma_{\mathrm{match},j}\!\cdot\!\mathrm{GSD}$,
where GSD denotes the ground sampling distance. The two components of
point $j$ share this standard deviation.
Each point contributes a $2\times p$ block $\bm B_j$ of $\bm B$. In our
implementation $\bm B=[\,\bm B_{\mathrm{poly}}\;\bm B_{\mathrm{line}}\,]$
is an \emph{empirical} basis---low-order polynomials in image
coordinates plus per-line terms---that spans the ground signatures of
inter-epoch orbit, attitude and scan-line offsets; we accordingly call
$\bm\xi$ revisit-bias coefficients and do not assign individual entries
an orbital meaning, which would require differentiating a physical
pushbroom model. The $2\times1$ blocks $\bm G_j$ propagate DEM error
through the difference of the two lines of sight,
$g_j=\tan\theta_{1,j}-\tan\theta_{0,j}$ resolved along the epipolar
direction (for a shared view this reduces to the $\Delta B/H$
view-angle factor). With a Gaussian DEM prior
$\bm e\sim\mathcal N(\bar{\bm e},\bm\Sigma_e)$, marginalizing $\bm e$
in \eqref{eq:obs} gives the working observations and covariance
\begin{equation}
\bm y=\Delta\bm u-\bm G\bar{\bm e},\qquad
\bm R=\bm\Sigma_n+\bm G\bm\Sigma_e\bm G^{T},
\label{eq:marg}
\end{equation}
so terrain leakage is carried in the noise model rather than dropped.
Attitude enters ground
units amplified by orbit altitude: $1''$ is $\approx2.4$\,m at
$H=500$\,km---the reason a satellite's pose knowledge is far poorer
than its deformation signal, and thus the regime in which a
differential parameterization gains the most.

\subsection{Informed differential estimator}
The revisit-bias coefficients $\bm\xi$ and displacements $\bm d$ are
solved jointly, on the marginalized observations of \eqref{eq:marg},
as a maximum-a-posteriori problem with Gaussian priors
$\bm\xi\sim\mathcal N(\bar{\bm\xi},\bm\Sigma_\xi)$ (nominal
georeferencing defines $\bar{\bm\xi}$; both parameter fields are
centered on their nominal values) and
$\bm d\sim\mathcal N(\bar{\bm d},\bm\Sigma_d)$,
$\bm\Sigma_d=\operatorname{diag}(\sigma_{d,j}^2)$ (a common $\sigma_d$
unless stated):
\begin{equation}
\begin{bmatrix}\widehat{\bm\xi}\\ \widehat{\bm d}\end{bmatrix}
=\arg\min_{\bm\xi,\bm d}\;
\big\|\bm y-\bm B\bm\xi-\bm d\big\|_{\bm R^{-1}}^{2}
+\big\|\bm\xi-\bar{\bm\xi}\big\|_{\bm\Sigma_\xi^{-1}}^{2}
+\big\|\bm d-\bar{\bm d}\big\|_{\bm\Sigma_d^{-1}}^{2},
\label{eq:map}
\end{equation}
where every weight is a precision (inverse-covariance) matrix, so DEM
error, matching noise, the geometric-bias prior and the displacement
prior close in a single probabilistic model. The Gaussian posterior
covariance is
\begin{equation}
\bm\Sigma_{\mathrm{post}}
=\big(\bm A^{T}\bm R^{-1}\bm A+\bm P^{-1}\big)^{-1},\quad
\bm A=[\bm B\;\;\bm I],\quad
\bm P=\operatorname{blkdiag}(\bm\Sigma_\xi,\bm\Sigma_d),
\label{eq:post}
\end{equation}
whose $\bm d$-block is the displacement posterior covariance.
The displacement prior is \emph{structurally exclusive}: it constrains the inter-epoch change itself, an unknown that exists only under the differential parameterization.
An absolute pipeline that estimates per-epoch positions has no state on which to place it; to accept it,
it would have to couple its two epochs, at which point it has become
the differential estimator. Conventional destriping is recovered as a
\emph{robust special case}: with the quadratic data term replaced by a
robust loss $\rho$,
\begin{equation}
\min_{\bm\xi,\bm d}\sum_j\sum_{k=1}^{2}
\rho\!\left(\frac{[\bm y_j-\bm B_j\bm\xi-\bm d_j]_k}{\sigma_{n,j}}\right)
+\big\|\bm\xi-\bar{\bm\xi}\big\|_{\bm\Sigma_\xi^{-1}}^{2}
+\big\|\bm d-\bar{\bm d}\big\|_{\bm\Sigma_d^{-1}}^{2},
\label{eq:robust}
\end{equation}
taking $\rho=|\cdot|$ and a hard stationarity constraint
$\bm d_{\mathcal S_\ell}=\bm 0$ on the stable pixels of line $\ell$
makes the per-line component of $\widehat{\bm\xi}$ exactly the
per-line median of standard destriping,
$\widehat\xi_\ell=\operatorname{median}_{j\in\mathcal S_\ell}[\bm y_j]_k$,
applied componentwise.
We use \eqref{eq:robust} as robust initialization and the quadratic
model \eqref{eq:map} for estimation and uncertainty.

\subsection{Immunity of per-orbit biases}
Per-orbit differences in orbit position and attitude are common to all
samples of an acquisition and therefore lie in the column space of
$B$; the estimator assigns them to $\bm\xi$, and they do not
contaminate $d$. This is \emph{absorption} immunity, to be distinguished from the
\emph{cancellation} immunity of shared calibration error in the
fixed-camera estimator \citep{pamidraft}: there the error is common to
both epochs and vanishes in the difference; here the per-orbit offset
differs between epochs and is instead estimated and assigned to
$\bm\xi$. 
This absorption provides immunity only under explicit identifiability
conditions, which we state precisely:
\begin{proposition}[Identifiability]
For the unregularized model with $\bm A=[\bm B\;\;\bm I]$,
$\operatorname{null}(\bm A)=\{[\bm a;\,-\bm B\bm a]:\bm a\in\mathbb R^{p}\}$:
any displacement component lying in $\operatorname{col}(\bm B)$ can be
traded against the revisit-bias coefficients. If a stable set
$\mathcal S$ is imposed through $\bm d_{\mathcal S}=\bm 0$, the
coefficients are identifiable from the data if and only if
$\operatorname{rank}(\bm B_{\mathcal S})=p$. With soft Gaussian priors
$\bm P$, the posterior is proper if
$\operatorname{null}(\bm A)\cap\operatorname{null}(\bm P^{-1})=\{\bm0\}$.
\end{proposition}
\noindent In practice $\bm\xi$ and $\bm d$ are separated by the
displacement prior and the stable-area reference
(without them the along-track mean is unidentifiable, as shown in
Section~\ref{sec:e5});
the extra parameters cost information (the informed estimator's floor
sits above the destriping ladder on a sparse scene, Table~\ref{tab:e5});
and any true displacement lying in the column space of $B$ is absorbed
as if it were offset---which is why the correction must be fitted on
stable ground only (Section~\ref{sec:duel}, Table~\ref{tab:h2h}). Within these
conditions, coarse orbit and attitude do not harm the differential
solution, only widen the prior.
Section~\ref{sec:sim} measures the residual after a $5$\,m/$10''$
injection at a ratio of $0.989$ to the un-injected case.

\subsection{Prior-screened along-track chain}
\label{sec:screen}
Model the along-track settlement chain as ring/segment values coupled
by seam observations and per-segment attitude, single-end anchored,
with the displacement prior $d\sim\mathcal N(0,\sigma_d^2)$. The
information matrix after eliminating attitude is a discrete Helmholtz
operator, and the interior variance saturates rather than accumulating:
\begin{equation}
v_\infty=\frac{\sigma_0^2}{2\sinh\theta_c},\qquad
\cosh\theta_c=1+\frac{\sigma_0^2}{2\sigma_d^2},\qquad
\ell_c=\frac{1}{\theta_c},
\label{eq:screen}
\end{equation}
with weak-prior limit $v_\infty\approx\sigma_0\sigma_d/2$ and
correlation length $\ell_c\approx\sigma_d/\sigma_0$ segments; as
$\sigma_d\to\infty$ the random walk is recovered. The derivation is
a tridiagonal-Toeplitz Green's-function argument
\citep{meurant1992}; \eqref{eq:screen} is its satellite
parameterization.

\subsection{Pushbroom: the revisit offset becomes $\xi(t)$}
\label{sec:pushbroom}
A frame camera's revisit offset is a constant vector; a pushbroom
line-array acquires each line at a different time, so its offset is a
per-line function $\xi(t)$. Parameterizing $\xi$ as a low-order
function of line number (or per-line with a smoothness prior) and
retaining the displacement prior restores the immunity of
Section~\ref{sec:screen} to attitude jitter: the identifiability of the
time-varying offset comes precisely from the displacement prior, which
is the estimation-theoretic reading of ``destriping the displacement
field.'' Section~\ref{sec:sim} shows a low-order offset leaks jitter in
full while the per-line offset holds the floor flat across jitter
grades.

\section{Accuracy Envelope by Simulation}
\label{sec:sim}

We characterize the estimator on a first-order single-component
pushbroom model (protocol: $T=40$ trials, bootstrap $2000$; $20\times20$
grid; $H=500$\,km; baseline orbit $5$\,cm, attitude $2''$, drift
$1''$/epoch, line jitter $0.2''$, DEM $10$\,m, match noise $0.1$\,px).
Five results define the envelope.

\subsection{E1. Immunity} Injecting $5$\,m orbit $+10''$ attitude
$+2''$ drift (over twenty meters of per-orbit ground bias) changes the
root-mean-square error (RMSE) by a ratio $0.989\ [0.972,1.002]$---statistically inseparable from
unity. Coarse orbit and attitude do not harm the differential solve.

\subsection{E2. Pushbroom $\xi(t)$} Under line jitter, a
constant\,+\,linear offset leaks the jitter in full (Table~\ref{tab:e2});
the per-line offset decouples the floor from jitter grade entirely,
holding $9.6$\,cm from $0.05''$ to $1''$ jitter. This is the numerical
confirmation of Section~\ref{sec:pushbroom}.

\begin{table}[t]
\centering
\caption{E2: line-jitter absorption by revisit-offset order
(along-track floor).}
\label{tab:e2}
\begin{tabular}{lccc}
\toprule
Line jitter & ground/line & low-order $\xi$ & per-line $\xi(t)$ \\
\midrule
$0.05''$ & 12\,cm & 17.3\,cm & \textbf{9.6\,cm} \\
$0.20''$ & 48\,cm & 64.0\,cm & \textbf{9.6\,cm} \\
$1.00''$ & 242\,cm & 318.4\,cm & \textbf{9.6\,cm} \\
\bottomrule
\end{tabular}
\end{table}

\subsection{E3. Control-free margin} 
Without ground control, an absolute two-epoch estimator with the same prior information has an RMSE of $5.85$\,m. Its pose is unidentifiable, so the attitude bias enters the epoch difference. The informed differential estimator has an RMSE of $9.6$\,cm, corresponding to a factor of $62$ improvement. 
Given eight stable anchor points to both estimators, the absolute route under a standard low-order RPC
correction still reads $79$--$81$\,cm against $8$--$16$\,cm for the
differential: a $5$--$10\times$ margin, CIs excluding unity, sourced in
the per-line jitter that eight anchors cannot pin.

\subsection{E4. Noise floor and DEM leakage} At $0.5$\,m GSD with
within-tube revisit ($\Delta B/H\!\approx\!5\times10^{-4}$) the
$2\sigma$ minimum detectable deformation is $9.4$\,cm per component at
a single point, and $0.7$\,cm when the two components of a $100$-point
patch are averaged jointly ($9.4/\sqrt{200}$; a single-component
$100$-point mean gives $0.9$\,cm); at $10$\,m GSD, $38$ and
$2.7$\,cm respectively. These patch figures assume independent
matching noise across points and components; spatially correlated
error and prior-induced shrinkage bias are not included and the
general contrast form is
$\mathrm{MDD}_{2\sigma}(\bm c)=2\sqrt{\bm c^{T}\bm\Sigma_{d\mid y}\bm c}$
with $\bm\Sigma_{d\mid y}$ the displacement block of
\eqref{eq:post}. The view-angle difference is the dominant design variable:
an agile $5^\circ$ cross-view pair leaks a $10$\,m DEM as
$87$\,cm/point through the $\Delta B/H\!\approx\!0.087$ lever, erasing
the GSD advantage. 
These results support the use of same-view image pairs. Cross-view
pairs require increasingly accurate elevation data as the difference
in viewing angle increases.

\subsection{E5. Engineering ladder and the observability boundary}
\label{sec:e5}
Against the standard optical ladder (Table~\ref{tab:e5}), naive
same-name differencing is dominated by per-orbit bias; a low-order
polynomial correction removes most of it; per-line destriping reaches
the match floor. The L2 informed estimator is statistically equivalent
to the destriping ladder on sparse deformation and superior on
dense/large-scale fields (Section~\ref{sec:real}); its added value is the
error budget, the uncertainty output, and the fusion interface, not a
single-pair accuracy miracle. The dense-field case in E5 marks the boundary: for a dense,
smooth along-track deformation, the field's along-track mean is
unidentifiable from attitude drift without control---both the ladder
and the differential estimator return the root-mean-square (RMS) signal amplitude as error. This is
an information boundary of the problem, to be reported, not an
estimator defect.

\begin{table}[t]
\centering
\caption{Comparison of the engineering baseline and the informed
differential estimator using along-track RMSE for the sparse-slip
scene.}
\label{tab:e5}
\begin{tabular}{lc}
\toprule
Method & RMSE (cm) \\
\midrule
same-name differencing & 584.6\, \\
$+$ low-order (RPC-style) correction & 64.6\, \\
$+$ per-line destriping (standard) & \textbf{5.2\,} \\
informed differential (per-line $\xi$, $\sigma_d$ prior) & 9.6\, \\
\bottomrule
\end{tabular}
\end{table}

The displacement-prior scale $\sigma_d$ has a two-sided optimum in the
chain regime: within $[0.5,2]\times$ the true magnitude the error is
flat; tightening to $0.1\times$ costs $2.45\times$; 
loosening it to $10\times$ the true magnitude increases the error by a factor of $1.7$, although it remains more accurate than the solution without a prior. These results suggest setting $\sigma_d$ within a factor of two of the expected displacement magnitude.

\section{Real-Data Validation}
\label{sec:real}

The three cases examine complementary operating conditions. The
Ridgecrest case contains extensive stable terrain and tests agreement
with conventional destriping. The Kahramanmara{\c{s}} case evaluates
the effect of cross-view geometry. The Baltoro case represents a
moving-majority scene with limited stable terrain and tests the
prior-constrained estimator. The controlled stress tests that isolate the mechanism follow in Section~\ref{sec:duel}.

\subsection{Ridgecrest 2019: 10\,m, coseismic, no control}
\label{sec:ridgecrest}
\begin{figure}[t]
\centering
\includegraphics[width=\textwidth]{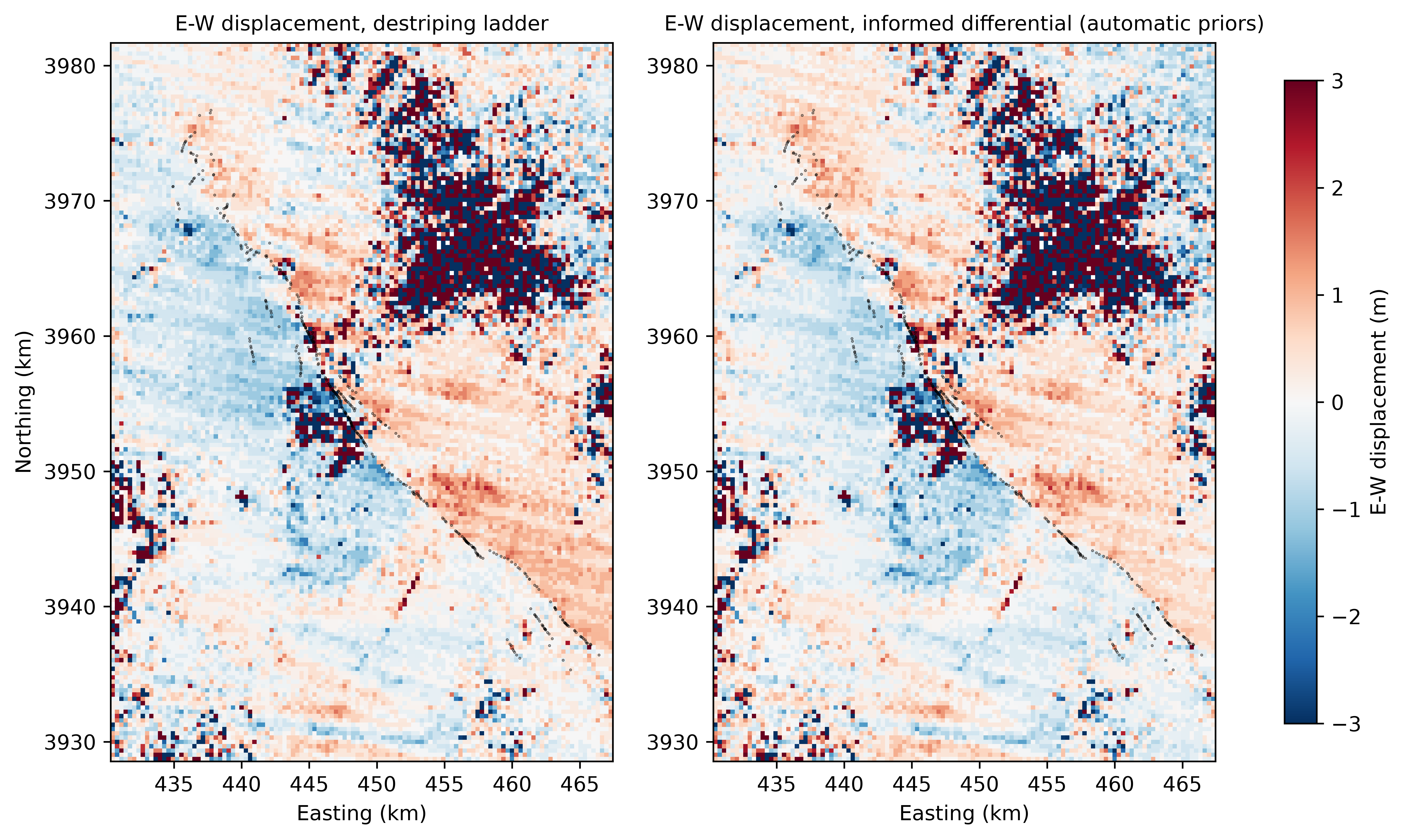}
\caption{Ridgecrest 2019 coseismic east--west displacement obtained
from the control-free Sentinel-2 pair. The left panel shows the
destriping baseline, and the right panel shows the informed
differential solution using the automatic prior rule described in
Section~\ref{sec:duel}. Black dots denote the USGS field observations
of surface rupture.
The lobed coseismic pattern and the discontinuity along the mapped
rupture are recovered in a single pass; the salt-flat decorrelation
zone (upper right) is flagged for masking.}
\label{fig:ridgecrest}
\end{figure}

The 2019 Ridgecrest sequence ($M_{\mathrm{w}}\,6.4$ and $M_{\mathrm{w}}\,7.1$) is imaged by Sentinel-2
(band B04, 10\,m) on 3 and 8 July 2019, using the same relative orbit
R027, over a $38\times54$\,km area. We match on a $64$/$32$ tile grid
with upsampled cross-correlation ($19{,}496$ valid tiles, $30$\,s) and
estimate the ladder and the informed differential, correcting on a
stable-area mask ($>8$\,km from the rupture, salt-flat decorrelation
zone removed); the resulting east--west displacement fields are shown
in Fig.~\ref{fig:ridgecrest}. Field truth is the USGS surface-displacement dataset (341 $M_{\mathrm{w}}\,7.1$ points).

\begin{figure}[t]
\centering
\includegraphics[width=0.6\textwidth]{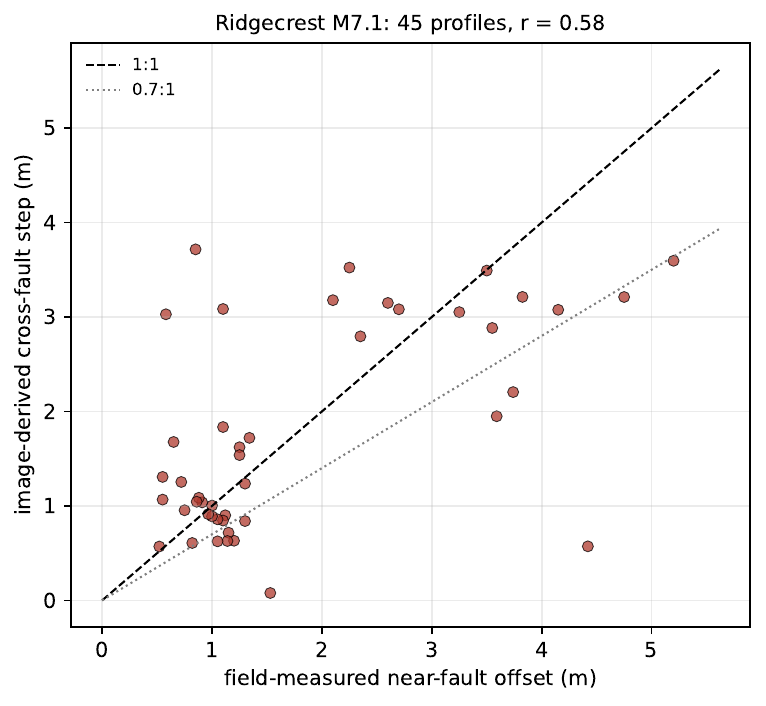}
\caption{Ridgecrest $M_{\mathrm{w}}\,7.1$: image-derived cross-fault step versus
field-measured near-fault offset for all 45 deduplicated profiles
with field offset $>0.5$\,m. The relation is unbiased (ratio
median 1.00); scatter grows toward the noise floor.}
\label{fig:profiles}
\end{figure}

\begin{figure}[t]
\centering
\includegraphics[width=0.85\textwidth]{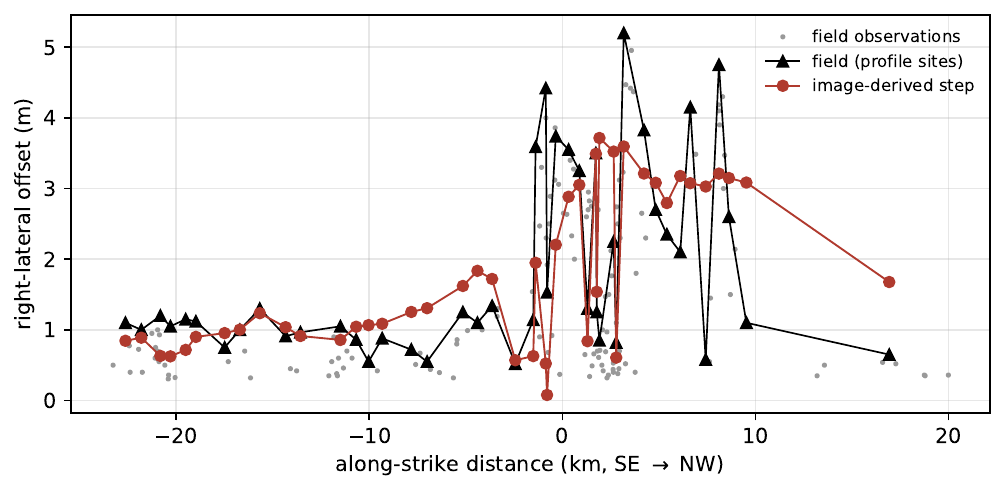}
\caption{Along-strike slip distribution: image-derived steps (red)
track the field measurements (black) over $45$\,km of rupture; the
image aperture smooths the near-fault concentration at the slip
maxima, the structure reported by field--image comparisons of this
event.}
\label{fig:slipdist}
\end{figure}

\begin{figure}[t]
\centering
\includegraphics[width=0.5\textwidth]{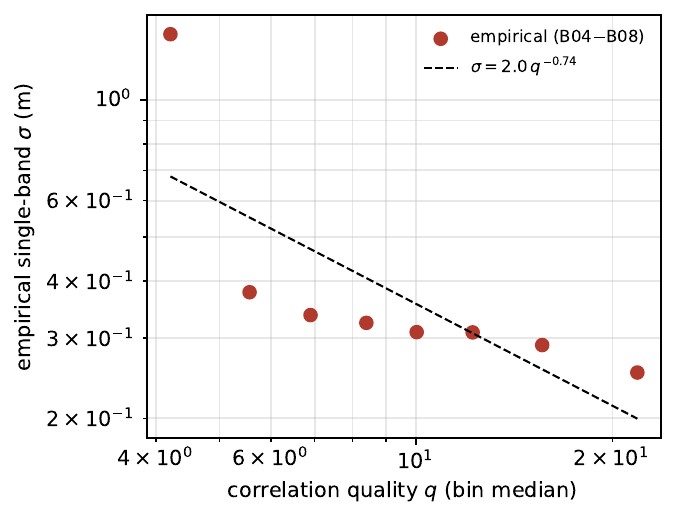}
\caption{Uncertainty calibration from two-band redundancy (Ridgecrest): empirical single-band $\sigma$ versus correlation quality $q$ (8 quantile bins) with a power-law summary; the shape is a plateau with a decorrelation knee.}
\label{fig:calib}
\end{figure}

The stable-area single-tile noise floor is $0.51$\,m for the
destriping ladder and $0.48$--$0.54$\,m ($0.05$\,pixel) for the
informed estimator under the automatic prior rule of
Section~\ref{sec:duel} (Supplementary Table~S8); the two are
statistically the same on this stable-ground-rich scene. The E4
envelope predicts $0.38$\,m at the 10\,m within-tube grade: the first
real pass lands within a factor $1.3$ of the pre-computed budget, with
a measured match noise ($0.05$\,px) better than the simulation's
conservative $0.1$\,px assumption. A deliberately tight displacement
prior ($\sigma_d=2$\,m) lowers the informed floor to $0.41$\,m but
absorbs part of the coseismic signal (cross-fault step $+2.5$ against
$+3.2$\,m)---the two-sided trade-off analyzed in
Section~\ref{sec:duel}, and the reason we report the automatic number. 
Along the rupture, cross-fault profiles at every field-measured offset above $0.5$\,m ($45$ profiles after
deduplication at a spacing of 500\,m; Fig.~\ref{fig:profiles}) give an unbiased image-to-field relation: ratio median $1.00$ (interquartile $[0.74,1.28]$), correlation $r=0.58$---moderate because half the
field offsets sit within a few multiples of the $0.5$\,m noise floor, where profile scatter is expected. An independent check against the continuous GNSS network is also available: on the full
$110$-km tile, the in-plane field at the $11$ Nevada Geodetic
Laboratory stations \citep{blewitt2018} (coseismic offsets of the two
events summed; $600$-m tile medians) differs from GNSS, with Gaussian-scaled median absolute
deviation ($\mathrm{MAD}_\sigma$)/RMS values of $0.14$/$0.31$\,m in East and $0.27$/$0.25$\,m in
North---consistent with the single-tile floor, and with a spatially
coherent North residual whose origin is identified in Section~\ref{sec:disc}. 
A three-scene closure test using the 3, 8, and 13 July 2019 Level-2A images further characterizes the error structure. All displacement fields are accumulated displacements (not velocities) resampled to the common tile grid of the first epoch, so the linear closure $d_{AB}+d_{BC}-d_{AC}$ applies; the finite-deformation composition $\bm d_{AC}(\bm x)=\bm d_{AB}(\bm x)+\bm d_{BC}(\bm x+\bm d_{AB}(\bm x))$ differs from it by a term of order $|\bm d|\,\|\nabla\bm d\|$, sub-millimetric for meter-level offsets on $600$-m tiles. Statistics are Gaussian-scaled median absolute deviations ($\mathrm{MAD}_\sigma$) of the fault-parallel component over all stable tiles: the closure residual has $\mathrm{MAD}_\sigma$ $0.46$\,m, compared with $0.77$\,m for a single pair.
Under mutually independent pair errors the expected ratio is $\sqrt{3}\approx1.7$ (the familiar $\sigma^2_{AC}=\sigma^2_{AB}+\sigma^2_{BC}$ requires independent one-dimensional components and no shared-scene covariance---precisely what closure tests); for errors associated entirely with individual scenes the ratio is zero.
Writing $\bm\epsilon_{ij}=\bm s_j-\bm s_i+\bm p_{ij}$ with scene terms $\bm s$ and pair terms $\bm p$, the scene terms cancel in the closure sum, and under independent, equal-variance components the scene fraction of pair-error variance is $f_{\mathrm{scene}}=1-\tfrac13(\sigma_c/\sigma_{\mathrm{pair}})^2$; the measured ratio of $0.60$ gives $f_{\mathrm{scene}}\approx88\%$.
(The held-out prior-audit protocol of Supplementary Table~S9 reports smaller $\mathrm{MAD}_\sigma$ values---closure $0.306$\,m, per-pair $0.26$--$0.45$\,m---because it evaluates the automatic-prior estimator on held-out halves of the stable set; the two protocols answer different questions and their numbers are not interchangeable.)
Multi-epoch differential chains can exploit this error component.
The post--post pair (07-08 vs.\ 07-13, tectonically null) doubles as a null experiment: its cross-fault step is $-0.14$\,m where the coseismic pairs read $+2.5$\,m under identical processing---the coseismic signal is real, and the null level sits at the noise floor. A band-level internal
check closes the loop without any truth: solving the pair
independently on B04 and B08, the two displacement fields agree to
a stable-area MAD of $0.45$--$0.49$\,m, consistent with two
largely independent realizations of the single-band floor. The same two-band redundancy provides a \emph{pair-specific calibration
of the matching uncertainty}---it does not by itself cover the
estimation uncertainty of $\bm\xi$, DEM-error correlation, stable-area
mis-selection, or prior-induced shrinkage, which enter through the full
posterior covariance \eqref{eq:post} and are audited at field level by
the closure and stable-area diagnostics rather than propagated
pointwise. Binning tiles by correlation quality $q$ and measuring the empirical single-band scatter per bin yields a monotone relation spanning a $5.5\times$ dynamic range (from $0.25$\,m at the best-texture bin to $1.39$\,m in the lowest, decorrelating bin; Fig.~\ref{fig:calib}). Its shape is a plateau ($0.25$--$0.38$\,m over the upper seven bins) with a decorrelation knee, not a clean power law: a fit $\sigma\approx2.0\,q^{-0.74}$ summarizes it with a $23\%$ median bin residual, and across four scene pairs (two Ridgecrest processing levels, Baltoro 2024 and 2025) fitted exponents range from $0.24$ to $0.65$ (Supplementary Fig.~S1). 
The result should therefore be interpreted as a calibration procedure rather than a universal relation. Two-band redundancy can be used to estimate $\sigma(q)$ separately for each multispectral image pair. Correlation-peak quality can then be converted into a per-point standard deviation instead of being used only as a masking threshold. 
The full along-strike comparison (Fig.~\ref{fig:slipdist}) closes the loop at the signal level.
The $34$\,m raw RMS is entirely the Searles salt flat, a
decorrelation zone flagged for masking.

\subsection{Kahramanmara{\c{s}} 2023: 0.52\,m, cross-view}
\label{sec:kahramanmaras}
\begin{figure}[t]
\centering
\includegraphics[width=\textwidth]{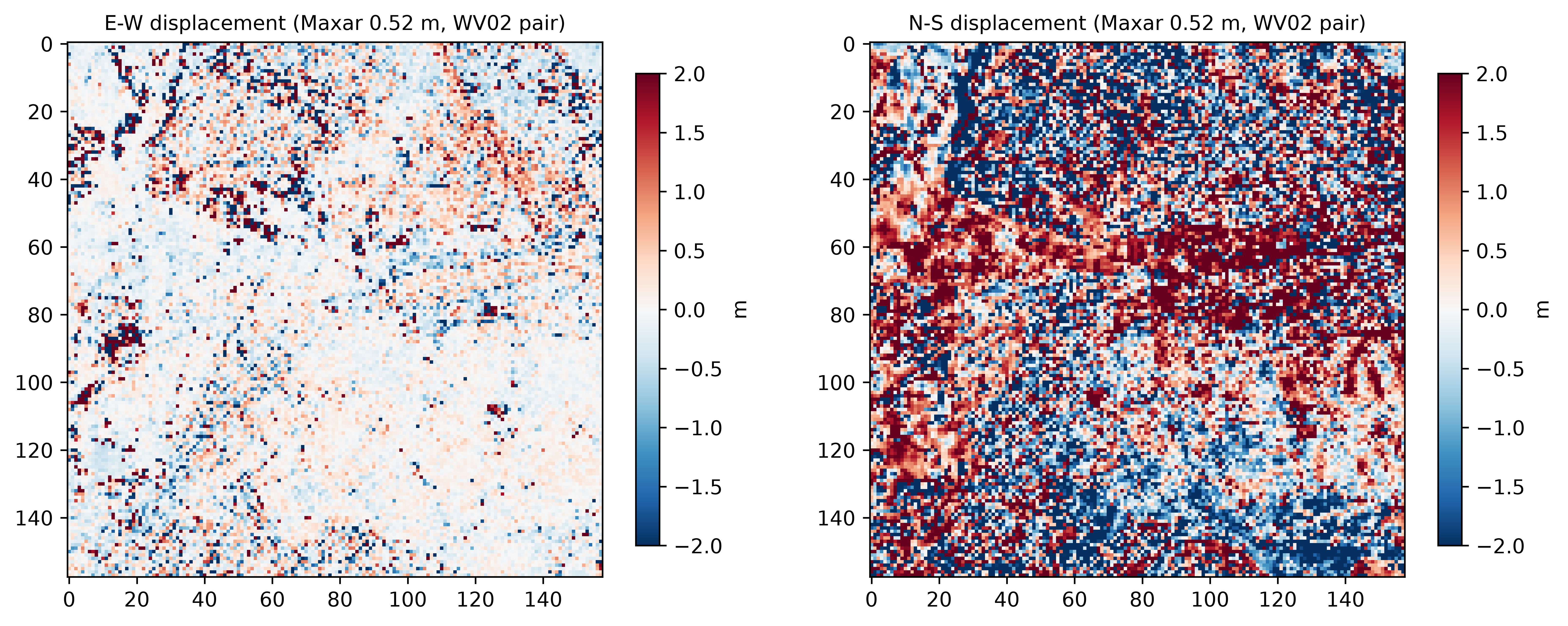}
\caption{Kahramanmara{\c{s}} 2023 cross-view WorldView-2 pair
(Narl{\i} segment, 0.52\,m): E-W (left) and N-S (right)
displacement after detrending. The strong anisotropy---quiet
across-look field versus noisy along-look field---is the
look-direction resolution-degradation penalty of cross-view
pairing, uncorrectable by any DEM (Section~\ref{sec:kahramanmaras}).}
\label{fig:maxar}
\end{figure}

The 2023 Kahramanmara{\c{s}} $M_{\mathrm{w}}\,7.8$ event over the Narl{\i} segment is imaged
by Maxar Open Data WorldView-2 panchromatic observations acquired on
2 January 2023 before the earthquake (off-nadir $21^\circ$, $0.52$\,m)
and on 12 February 2023 after the earthquake (off-nadir $34^\circ$,
$0.60$\,m, resampled to the $0.52$\,m grid). This is a deliberately
adverse \emph{cross-view} pair (Fig.~\ref{fig:maxar}). The stable-area floor is strongly
anisotropic (Table~\ref{tab:maxar}): $0.27$\,pixel across the
look-direction difference, $1.73$\,pixel along it. The across-look
direction sits at the sub-meter noise floor the envelope predicts---
the sub-meter grade's first field point. 
We investigated the excess along-look error using regression against the Copernicus GLO-30 DEM and terrain-correlation analysis. The results show that DEM leakage is not its dominant source. 
Height regression removes only $2\%$ of it (transfer coefficient $0.0075$\,m/m against a $\Delta\tan=0.29$ potential); 
the residual is uncorrelated with elevation, high-passed elevation, and slope ($|r|\le0.07$); and its spatial autocorrelation is near-white. 
The bias-type DEM term has already been absorbed by the analysis-ready data (ARD) product's own surface-model orthorectification; 
what remains is a look-direction \emph{resolution degradation}---the $34^\circ$ off-nadir acquisition
stretches the ground sample along the line of sight, compounded by
the $0.60\!\to\!0.52$\,m resampling---which is noise-type and
uncorrectable by any DEM. The design consequence is stronger than
the one E4 models: a cross-view pair pays the DEM-leakage term
\emph{and} a look-direction noise penalty ($6\times$ here) that
survives even a perfect DEM. 
Same-view pairing is therefore recommended for sub-meter deformation measurement. 
A same-view control pair confirms this reading from the other side: a WorldView-3 pair
with only $3.2^\circ$/$4.9^\circ$ view difference, acquired on
17 October 2022 and 21 February 2023 over stable terrain far west of the rupture, restores full
isotropy---$0.47$\,px on both axes, anisotropy ratio $1.02$ against
the cross-view pair's $6.4$---at a floor elevated by its 127-day
cross-season gap rather than by geometry. The best observed
sub-meter floor therefore remains the $0.27$\,px of the short-gap
pair's across-look axis, and the two pairs bracket the design
space: view difference sets the anisotropy, temporal gap sets the
isotropic floor.

\begin{table}[t]
\centering
\caption{Kahramanmara{\c{s}} cross-view pair: anisotropic noise floor
(stable-area robust MAD $\sigma$).}
\label{tab:maxar}
\begin{tabular}{lcc}
\toprule
Direction & detrended & informed \\
\midrule
across view-angle diff. & $0.29$\,m ($0.55$\,px) &
\textbf{$0.14$\,m ($0.27$\,px)} \\
along view-angle diff. & $1.92$\,m ($3.69$\,px) &
$0.90$\,m ($1.73$\,px) \\
\bottomrule
\end{tabular}
\end{table}

\subsection{Baltoro glacier 2024: 10\,m, flow regime, three-scene triplet}
\label{sec:glacier}

\begin{figure}[t]
\centering
\includegraphics[width=\textwidth]{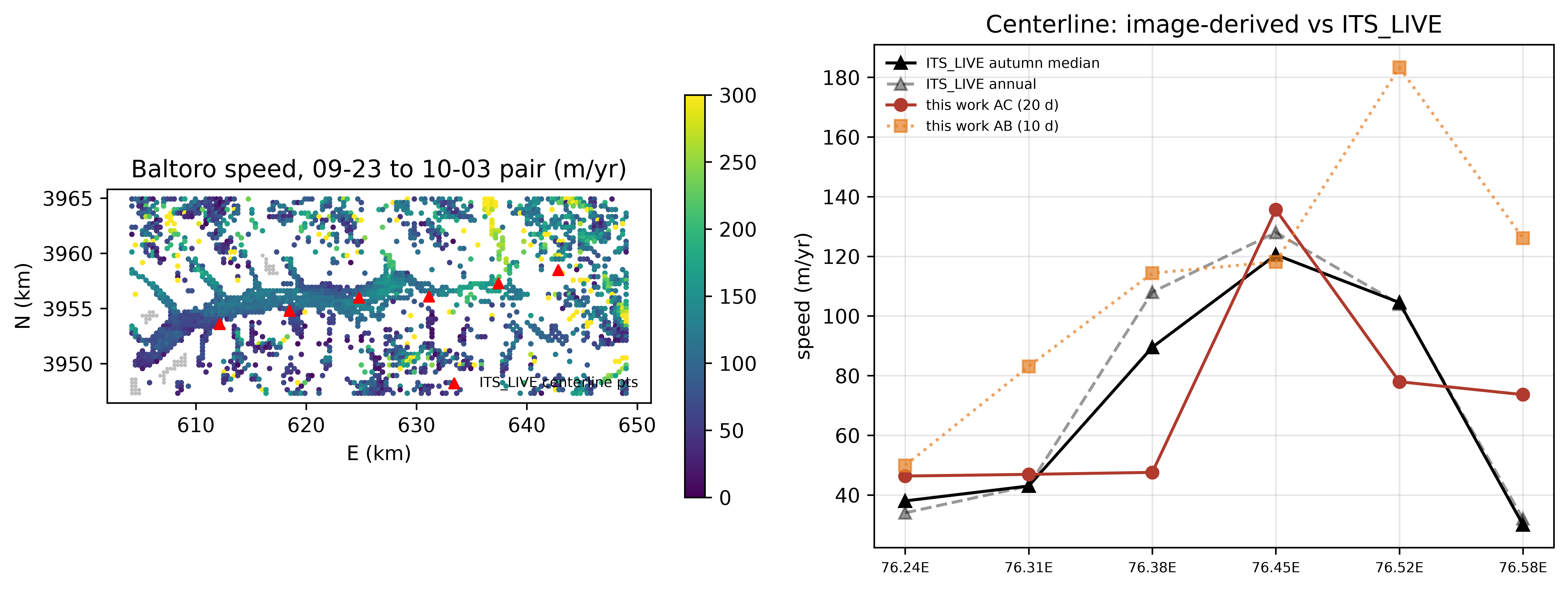}
\caption{Baltoro glacier, Sentinel-2A 10\,m, control-free. Left: speed
field from the 23 September--3 October pair (m\,yr$^{-1}$; gray = bedrock stable
area; triangles = ITS\_LIVE centerline points). Right: centerline
speed, this work (10-day and 20-day pairs) versus ITS\_LIVE autumn and
annual medians; the 20-day pair includes the fresh-snow scene.}
\label{fig:glacier}
\end{figure}

Glacier flow is the deformation regime where optical correlation is
most used and where a per-point uncertainty is most needed
\citep{dehecq2019,millan2022}; Baltoro is a debris-covered trunk of
the type whose climate response is known to be spatially variable
\citep{scherler2011}. We processed three cloud-free Sentinel-2A
same-orbit acquisitions over the Baltoro glacier (Karakoram; tile
43SFV, band~8, 23 September, 3 October, and 13 October 2024,
$45\times18$\,km AOI)
with the pipeline of Section~\ref{sec:ridgecrest} unchanged, with the
stable area taken as bedrock more than 1\,km outside the Randolph Glacier Inventory (RGI)
glacier outlines and the ITS\_LIVE v2 velocity product
\citep{lei2021} as reference (autumn-season median of all image pairs at
six centerline points). Two facts appear (Fig.~\ref{fig:glacier}).

First, the envelope holds in the flow regime. The 10-day pair's
bedrock floor is $0.47$\,m ($0.047$\,pixel) in the across-track
component---the same figure as Ridgecrest's $0.05$---and its
centerline speeds track the ITS\_LIVE autumn median point by point
from the terminus to the trunk maximum ($50\to118$\,m\,yr$^{-1}$
against $38\to120$; median relative difference $15\%$ over the trunk),
with the trunk's spatial structure---a continuous $100$--$200$\,m\,yr$^{-1}$
band, tributary inflows, terminal deceleration---resolved in a single
pass. The three-scene closure $d_{AB}+d_{BC}-d_{AC}$ gives ratios of
$0.79$ (E) and $0.47$ (N) on bedrock and $0.47$--$0.50$ on ice, in the
same range as Ridgecrest's $0.60$: the per-scene error cancellation
that the closure law predicts survives the transition from a static
scene to a flowing one. 
The same closure behavior is observed independently in the ITS\_LIVE
pair archive.
Because ITS\_LIVE reports velocities, each pair is first converted to an
accumulated displacement over its own time span
($\Delta t\,\bm v$; the closure identity for velocities is
$\Delta t_{AC}\bm v_{AC}=\Delta t_{AB}\bm v_{AB}+\Delta t_{BC}\bm v_{BC}$),
and all pairs of a triplet are evaluated at the same archive grid
points, so the sums compare like with like.
Applying the triplet statistic at the six centerline points ($15\,369$ short-baseline pairs,
$25\,354$ closed triplets) gives closure ratios of $0.61$ in both
components---indistinguishable from our $0.60$ at Ridgecrest---so about
$88\%$ of the pair-error variance in an operational velocity product
is per-scene and cancels in closure. The closure statistic is thus a
scene-quality flag that any pair product could carry, and the
per-scene decomposition is a property of repeat-pass optical
correlation, not of one pipeline.

Second, the triplet delivers the envelope's first measured \emph{decorrelation} sample. 
The 13 October scene contains approximately twice the bright-snow area of the other two scenes ($19\%$ compared with $9\%$), and its median radiance is $40\%$ higher. 
This early-October snowfall partially altered the surface texture. 
Every pair containing this scene has a noise floor approximately three times higher ($1.3$--$1.9$\,m) and biased accumulation-zone speeds. 
The 10-day pair that excludes it is therefore the primary result and the 20-day pair is retained for closure only. 
The lesson is a design rule the envelope did not contain: the temporal baseline must avoid snowfall events, a condition that is checkable from the scene statistics before correlation. 
We record as well what the single-pair result cannot yet do: at $1$--$3$\,m of ten-day displacement against a $0.5$--$1$\,m floor the signal-to-noise ratio is $2$--$6$, sufficient for the trunk but not for tributaries or the accumulation basin, which need a same-season interannual baseline or multi-pair stacking---precisely the multi-epoch fusion the differential model is built to accept.

\begin{figure}[t]
\centering
\includegraphics[width=\textwidth]{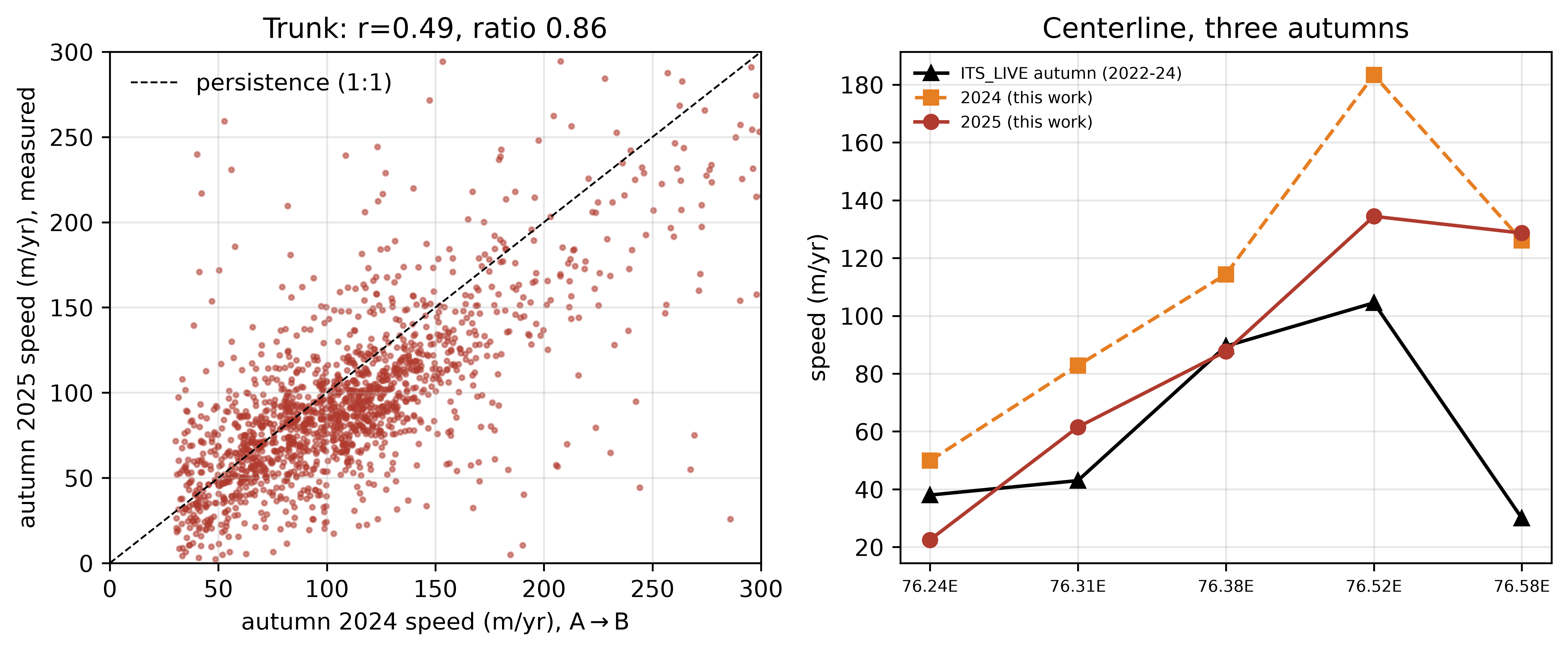}
\caption{Forward test. Left: autumn-2025 trunk speed measured from the
13--23 September 2025 pair versus the autumn-2024 field taken as a
persistence forecast (dashed 1:1). Right: centerline speed for the
ITS\_LIVE 2022--2024 autumn median, 2024 and 2025 from this work.}
\label{fig:glacier2025}
\end{figure}

Third, the product is testable forward in time. Repeating the
pipeline on the same-season 2025 triplet (Sentinel-2B, 13 September,
23 September, and 13 October 2025, read on the same grid) reproduces the 10-day bedrock
floor ($0.52$/$0.91$\,m E/N against $0.47$/$0.99$ in 2024) and the
closure ratio ($0.38$/$0.60$), and---strikingly---the same mid-October
snowfall ($22\%$ bright-snow area against $8\%$ a fortnight earlier),
which fixes the tasking rule as seasonal rather than incidental. Taking
the 2024 speed field as a persistence forecast of 2025
(Fig.~\ref{fig:glacier2025}), the 1608 trunk tiles scatter about the
1:1 line with a speed-difference MAD of $28$\,m\,yr$^{-1}$, below the
$37$\,m\,yr$^{-1}$ that the two floors predict for pure noise: the
forecast holds to within measurement noise. The median ratio of $0.86$,
consistent across three along-flow segments ($0.86$, $0.82$, $0.88$),
is a candidate interannual slowdown of the order reported for the
region \citep{dehecq2019}; we do not claim it, because a shared
systematic of that size cannot yet be excluded from two seasons, and a
third season is the honest test. What can be claimed is the
instrument's role: the interannual difference now comes with a
per-point posterior uncertainty and a closure check, which is what turns a
velocity mosaic into an observation a mass-balance model can ingest. An attempt to resolve intra-seasonal dynamics from all 5--15-day
pairs of seven autumns failed for an instructive reason---the stable
area of this AOI is too small to pin the correction to better than a
ten-day displacement, and the closure statistic flags it
(Supplementary Section~S2); we make no claim about intra-seasonal
deceleration.

The three cases cover the envelope's ends: two resolution classes
($10$\,m and sub-meter), two deformation regimes (coseismic step
and glacier flow), same-view and cross-view, control-free throughout. The noise floor is verified at both grades; the
same-view-pairing design rule is verified---and sharpened with a
second, noise-type penalty term---on the cross-view case; the
budget-vs-measurement agreement is demonstrated on the same-view
case.

\section{Limits of destriping and benefits of the displacement prior}
\label{sec:duel}
Sections~\ref{sec:sim} and~\ref{sec:real} establish the envelope and show that on a single pair
with ample stable ground the differential estimator and per-line
destriping agree. This section asks the operational question directly,
on the real offset fields already computed: what decides the signal
(the correlator or the post-processing), what happens when the stable
area is thinned or is naturally scarce, what a known truth says, and
whether the priors that make the difference can be set without a hand.

\subsection{Kernel versus post-processing}
A controlled comparison separates the effects of the matching kernel and the subsequent geometric correction (Table~\ref{tab:h2h}):
on the same scenes, stable areas and criteria, three matching kernels (frequency phase correlation as in COSI-Corr; an autoRIFT-style normalized cross-correlation reimplemented here; and the autoRIFT package itself, run on identical $64$-px chips) move the noise floor within a factor of $1.3$ of each other---the autoRIFT package, with its
default high-pass prefilter switched off, is the best matcher at
$0.038$/$0.050$\,px against our $0.049$/$0.073$---whereas the
post-processing decides the signal by a factor of two. (With the
prefilter on, autoRIFT reports a $0.024$-px floor but the coseismic
step falls by $7\%$ and the glacier trunk median by $39\%$; the
smoothing is inside the package chain, so we report the setting that
preserves the signal.)
The three kernels compared here are all window-correlation estimators,
whose displacement transfer characteristic is set by the chip size;
the interchangeability reported above should therefore not be
extrapolated to matchers that carry a strong spatial smoothness prior,
such as learned dense optical-flow networks, whose response to
short-wavelength displacement may be strongly attenuated. For such
front ends the transfer characteristic has to be calibrated against
the spatial scale of the target signal before the floor can be read as
a precision.
On the static Ridgecrest scene the classical full-field
detrend-and-destripe and the masked differential correction agree to
$0.01$\,m, as the special-case theorem says they must; on Baltoro,
where $57\%$ of the area moves, the full-field heuristic absorbs half
the glacier flow into its ``correction'' ($58$ against
$106$\,m\,yr$^{-1}$ trunk median with the same kernel), while the
masked estimator recovers the ITS\_LIVE-consistent value. The gain is
not a better correlator; it is knowing what may be fitted to what.

\begin{table}[t]
\centering
\caption{Kernel $\times$ post-processing on identical inputs (10\,m,
$64$-px chips, $32$-px step). Floor: stable-area MAD$\sigma$. NCC is
an autoRIFT-style reimplementation; autoRIFT is the package
(conda-forge 2.1.1), shown with its default high-pass prefilter off
(signal-preserving) and on (default). Floors are in m; the signal is
the cross-fault step in m (Ridgecrest) or the trunk median speed in
m\,yr$^{-1}$ (Baltoro); ``heuristic'' is the full-field
detrend-and-destripe, ``masked'' the stable-area differential
correction.}
\label{tab:h2h}
\footnotesize\setlength{\tabcolsep}{4.5pt}
\begin{tabular}{llcccc}
\toprule
& & \multicolumn{2}{c}{Noise floor} & \multicolumn{2}{c}{Signal} \\
\cmidrule(lr){3-4}\cmidrule(lr){5-6}
Scene & Kernel & heuristic & masked & heuristic & masked \\
\midrule
Ridgecrest & PC  & $0.50$\,m & $0.49$\,m & $3.83$ & $3.86$ \\
Ridgecrest & NCC & $0.55$\,m & $0.53$\,m & $4.14$ & $4.09$ \\
Ridgecrest & autoRIFT (pf.\ off) & $0.39$\,m & $0.38$\,m & $4.73$ & $4.74$ \\
Ridgecrest & autoRIFT (default) & $0.28$\,m & $0.24$\,m & $3.56$ & $3.58$ \\
Baltoro    & PC  & $0.81$\,m & $0.73$\,m & $58$ & $106$ \\
Baltoro    & NCC & $0.94$\,m & $0.73$\,m & $63$ & $119$ \\
Baltoro    & autoRIFT (pf.\ off) & $0.58$\,m & $0.50$\,m & $62$ & $94$ \\
Baltoro    & autoRIFT (default) & $0.29$\,m & $0.33$\,m & $47$ & $65$ \\
\bottomrule
\end{tabular}
\end{table}

\subsection{Thinning the stable area: explain, then improve}
\begin{figure}[t]
\centering
\includegraphics[width=\textwidth]{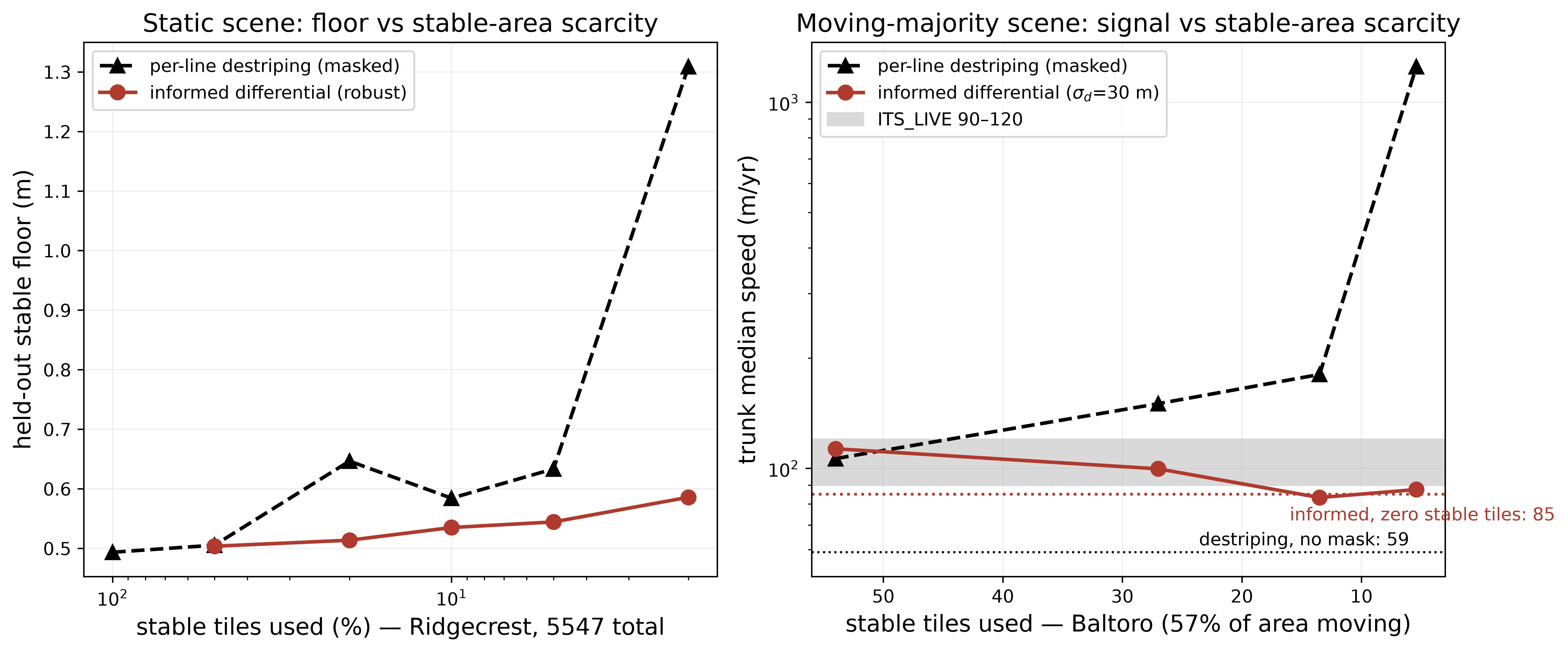}
\caption{Thinning the stable area. Left: Ridgecrest, held-out stable
floor as the fitting set shrinks from $50\%$ to $2\%$ of $5547$ tiles.
Right: Baltoro, trunk median speed as the $54$ bedrock tiles are reduced
to $5$; horizontal lines mark the zero-stable-tile informed solution
and full-field destriping.}
\label{fig:duel}
\end{figure}

Because per-line destriping is universal practice, the useful question
is not whether the differential model explains it but where the prior
makes a difference. We answer it by thinning the stable area
(Fig.~\ref{fig:duel}): the same tiles, the same strip geometry, the
stable set subsampled from $100\%$ to $2\%$ ($10$--$12$ random draws),
masked destriping against the informed estimator with per-tile priors
(stable tiles $\sigma_d=0.05$\,m as spatial reference; other tiles the
expected-displacement scale; strip offsets bounded by the attitude
jitter amplitude, $0.5$\,m). On the static Ridgecrest scene the
destriping floor rises from $0.49$ to $1.31$\,m as strips lose their
stable tiles, while the informed floor stays at $0.51$\,m down to $2\%$
with equal step fidelity. On Baltoro, where $57\%$ of the area moves
and only $54$ bedrock tiles exist, destriping produces a plausible estimate when all $54$ tiles are used
($106$\,m\,yr$^{-1}$), but becomes unstable as the number of stable
tiles decreases ($150$, $181$, $1251$ at $27$, $13$, $5$ tiles: strips without stable
tiles receive no correction), whereas the informed estimator degrades
gracefully ($102$, $94$, $84$, $87$) and still returns $86$ with
\emph{no} stable tile at all, from the statistical prior alone, against
an ITS\_LIVE reference of $90$--$120$; full-field destriping without a
mask returns $59$. Two honest boundaries come with this: the prior
scales must match the regime (setting the free-tile $\sigma_d$ to
$4$\,m instead of $20$\,m on the glacier pulled the trunk speed to
$62$--$85$---the plane absorbed flow---and the error was visible),
and robust re-weighting helps when moving tiles are a minority
(Ridgecrest) but hurts when they are the majority (Baltoro,
$90\to68$). Within those rules the advantage is structural: every
strip and every tile has a solution, degradation is smooth, and
operation continues where the heuristic has no stable ground to fit.

\subsection{Controlled truth: injected deformation on a real background}
Real data give no truth for the comparison above, so we inject known
deformation into the real Ridgecrest stable-area offsets (5094 tiles,
real noise and real stripes, $0.46$\,m background MAD) and measure the
recovery RMSE of both estimators (Table~\ref{tab:inject}; the
informed estimator with the same priors as above, unknown-tile
$\sigma_d=20$\,m, robust weighting on since the moving tiles are a
minority). Three shapes probe the three regimes. A local
$5$\,km/2\,m patch---the ordinary case---is recovered slightly better
by the informed estimator ($0.49$--$0.71$ vs.\ $0.65$--$0.69$\,m). A
smooth field spanning the whole area leaves only $\sim$$20$--$40$
truly stable tiles: destriping recovers it with all of them
($0.79$\,m) but breaks when they are thinned ($2.5$--$2.8$\,m), while
the informed estimator holds $1.19$\,m throughout---the graceful
degradation of the glacier case reproduced under known truth. A
strip-aligned band of deformation, finally, is the adversarial case
the theory predicts (Section~\ref{sec:e5}): it lies in the column space of the
revisit offset, the informed estimator with a $20$\,m unknown-tile
prior absorbs part of it ($1.14$--$1.30$ vs.\ $0.74$--$0.80$\,m), and
only an uninformative prior ($\ge50$\,m) removes the loss
(Supplementary Table~S6)---at which point, however, the real Baltoro
plane loses the regularization that kept it stable with $5$ tiles.
The prior scale therefore involves a two-sided trade-off. 
A prior that is too restrictive absorbs deformation resembling the sensor error, whereas an excessively broad prior provides insufficient regularization when the spatial reference is sparse. 
The rule we adopt and recommend---unknown tiles at roughly ten times the expected displacement, stable tiles pinned, strip offsets bounded by the measured jitter amplitude ($0.5$\,m here), robust weighting only when moving tiles are a
minority---is the setting behind every prior-constrained result of this study, is implemented as an automatic procedure below, and the sensitivity sweep is reported in full in the Supplement.

\begin{table}[t]\centering
\caption{Injected deformation on the real Ridgecrest background:
recovery RMSE (m) against known truth. Stable fraction = share of the
truly stable tiles made available for fitting; 8 random draws where
$<100\%$.}\label{tab:inject}
\footnotesize\setlength{\tabcolsep}{4.5pt}
\begin{tabular}{llcccc}\toprule
& & \multicolumn{2}{c}{destriping} & \multicolumn{2}{c}{informed} \\
\cmidrule(lr){3-4}\cmidrule(lr){5-6}
Injected field & stable & masked & full-field & plain & robust \\\midrule
local patch, $2$\,m, $\sigma\,5$\,km & 100\% & 0.65 & 0.86 & 0.49 & 0.49 \\
 & 20\% & 0.67 & 0.86 & 0.66 & 0.66 \\
 & 5\% & 0.69 & 0.86 & 0.71 & 0.70 \\
smooth field, $\pm2$\,m & 100\% (37 tiles) & 0.79 & 1.09 & 1.19 & 1.20 \\
 & 20 tiles & 2.47 & 1.10 & 1.19 & 1.20 \\
strip-aligned band, $1.5$\,m & 100\% & 0.80 & 1.50 & 1.16 & 1.14 \\
 & 5\% & 0.74 & 1.45 & 1.30 & 1.28 \\\bottomrule
\end{tabular}\end{table}

\subsection{Natural cut-outs: two seasons, two scenes}
\begin{figure}[t]\centering
\includegraphics[width=\textwidth]{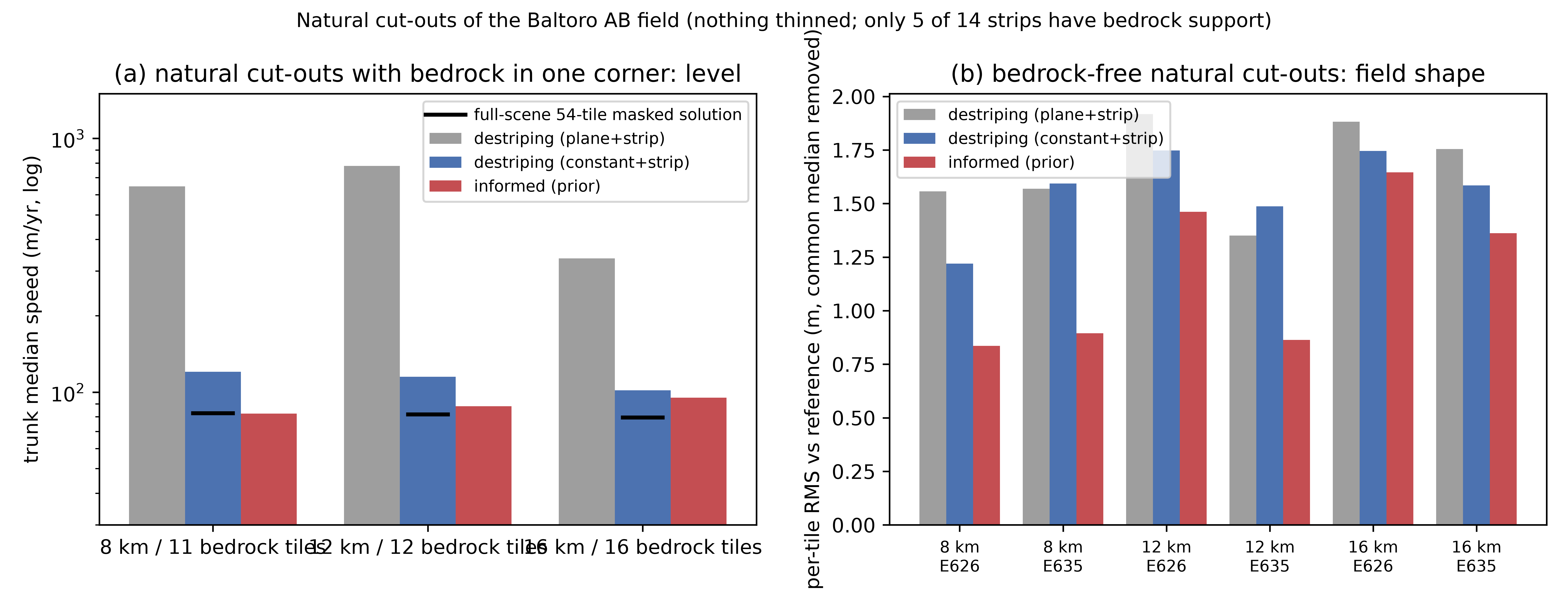}
\caption{Natural cut-outs of the Baltoro AB field, nothing thinned.
(a) Windows with bedrock in one corner: trunk median speed (log) for
plane-plus-strip destriping, constant-plus-strip destriping and the
informed estimator; black tick = full-scene 54-tile masked solution.
(b) Bedrock-free windows: per-tile RMS of the field shape (common
median removed) against the full-scene solution.}\label{fig:natural}
\end{figure}

The thinning experiment removes stable tiles by hand; the fair
question is whether the two failure regimes occur \emph{naturally}.
They do, on the very scene already analyzed
(Fig.~\ref{fig:natural}). First, of the 14 along-track strips that
cross the Baltoro area, only 5 contain three or more bedrock
tiles---the other 9 are inherently uncorrectable by masked destriping,
whatever the analyst does. Second, we cut the real AB offset field into
$8$, $12$ and $16$\,km windows centered on the trunk---the footprint of
a commercial ARD tile or of any glacier-scale subset---and process
each window on its own, changing nothing but the cut-out. Three
windows contain bedrock only in one corner ($11$--$16$ tiles); six
contain none. In the corner-bedrock windows plane-plus-strip
destriping extrapolates the plane through the clustered tiles and
returns $337$--$778$\,m\,yr$^{-1}$ for a trunk whose full-scene
reference is $80$--$83$; the more conservative constant-plus-strip
variant returns $102$--$120$ ($+25$--$45\%$); the informed estimator,
with its polynomial prior scaled to the window
($10$\,m$\times L/45$\,km, i.e.\ the measured plane amplitude),
returns $82$--$95$. In the six bedrock-free windows every method loses
the absolute level (the raw pair carries a $2.9$\,m co-registration
offset, so only the field \emph{shape} is testable): destriping falls
back to a full-field fit that subtracts flow, and its per-tile RMS
against the full-scene solution is $1.2$--$1.9$\,m, while the informed
estimator's is $0.8$--$1.7$\,m, better in six of six windows. 
Two limitations should also be noted: in the corner-bedrock windows the constant variant's shape RMS ($1.2$--$1.9$\,m) is slightly better than the informed estimator's ($1.6$--$2.4$\,m) even though its level is worse; 
and the informed level in bedrock-free windows is set by an implicit ``field median is zero'' assumption, which is wrong on all-ice cut-outs for any method. 
The scenario of this section is therefore not a synthetic stress test but the ordinary situation of
a glacier tile.
The same holds a season later and on a different kind of scene. The
2025 Baltoro pair (51 bedrock tiles; 4 of 14 strips supported) gives,
on the five bedrock-free windows, shape RMS $0.34$--$0.83$\,m for the
automatic-prior estimator against $0.85$--$1.22$\,m for either
destriping variant---five of five, by a factor $2$--$3$---and on the
four corner-bedrock windows ($4$--$24$ tiles) trunk speeds within
$13\%$ of the full-scene reference against $+8$--$38\%$ for
constant-plus-strip and $122$--$776$\,m\,yr$^{-1}$ for
plane-plus-strip. On Ridgecrest, $8$--$12$\,km windows centered on the
rupture contain no far field at all: the estimator recovers the
full-scene cross-fault step ($3.75$\,m) exactly with a shape RMS of
$0.23$--$0.38$\,m, where destriping falls back to a full-field fit and
returns $3.15$--$3.39$\,m with $0.42$--$0.92$\,m; at $16$--$24$\,km,
when hundreds of far-field tiles enter the window, destriping is
equal or better ($0.37$ against $1.00$\,m at $24$\,km)---the first
act again, and the boundary of the claim (Supplementary Table~S7).

\subsection{Priors by rule, not by hand}
Every prior-constrained result above uses priors set by the rule
stated at the end of the previous paragraph. To remove the residual
suspicion that the numbers depend on per-scene hand-setting, we
implemented the rule as an automatic, empirical-Bayes procedure
(\texttt{auto\_priors} in the released code) that reads nothing but
the offset field: $\sigma_n$ from the MAD of nearest-neighbor
differences; the trend and strip prior scales from the RMS of a
MAD-clipped whole-field trend fit and of the per-strip medians (the
sample variance of each effect is its prior variance); the
unknown-tile scale as ten times the 90th percentile of the residual
field; stable tiles pinned at $0.05$\,m; robust weighting on when the
moving fraction---from a physical mask such as the RGI outline where
one exists, else from the residual---is below one half. Two earlier
versions of the rule estimated the trend scale on the stable tiles
and failed on cut-outs where those tiles sit in one corner (their
extrapolated plane inflated $\sigma_{\rm poly}$ to $6$--$15$\,m); the
released version therefore never uses the spatial distribution of the
stable tiles for the scales, only as pins. Supplementary Table~S8
re-runs every prior-constrained experiment of this paper with the
automatic rule. The hand-set and automatic numbers agree to
$0.05$--$0.2$\,m in the controlled-truth tests, to $\le10\%$ in trunk
speed on the natural cut-outs, and the automatic rule keeps the
Ridgecrest floor flat ($0.50$--$0.57$\,m) where destriping rises to
$1.31$\,m. 
Two limitations remain. 
First, on the Ridgecrest headline pair the automatic rule gives an informed floor of $0.48$--$0.54$\,m, \emph{statistically the same as} the destriping ladder ($0.51$--$0.56$\,m); 
the $0.41$\,m we obtained with a hand-set $\sigma_d=2$\,m prior is bought by absorbing signal (cross-fault step $+2.5$ against $+3.2$\,m for both destriping and the automatic rule)---the two-sided trade-off of the previous paragraph, seen on real data, and the reason Section~\ref{sec:ridgecrest} now reports the automatic number. 
Second, the robust switch is the one residual judgement: Baltoro's valid tiles are $38\%$ glacier, the rule therefore enables Huber weighting, and Huber costs $\sim$$20\%$ of the trunk speed at five stable tiles ($70$ against $87$ with L2)---still eighteen times better than destriping's $1251$; we report both columns.

\paragraph{Identifiability limits and the strict-hybrid estimator}
A natural expectation is that an estimator containing destriping as a special case should never do worse than it. 
That holds on the identifiable part of the problem---strips with stable support, and deformation outside the column space of $\bm B$---and it is what the thinning, injection and cut-out experiments show. 
It cannot hold on the unidentifiable part: on a strip with no stable tile, deformation aligned with the strip is indistinguishable from the strip's revisit offset, and the outcome is decided by the prior alone.
Destriping's implicit prior there is $\xi=0$ (no correction); the informed estimator's is $\xi\sim\mathcal N(0,\sigma_\xi^2)$ (partial
absorption). Neither is right in general: the injected strip-aligned
band favors the former ($0.80$ against $1.33$\,m), the real Baltoro
strips favor the latter ($87$ against $1251$\,m\,yr$^{-1}$). 
Because neither assumption is universally valid on the unidentifiable subspace, we also evaluate an alternative member of the estimator family: 
a \emph{strict hybrid} that applies masked destriping exactly wherever it is
determined (median of the stable tiles on every supported strip; a plane only when the stable tiles span the window) and the prior only on the unsupported remainder. 
It reproduces destriping to the last digit where destriping works (Ridgecrest floor $0.51/0.65/0.58/0.63$\,m at
$50/20/10/5\%$ stable, identical to destriping) and lifts it where
destriping fails (Baltoro trunk $113/100/92$ at $27/13/5$ tiles against
$150/181/1251$; injected band $0.73$--$0.77$ against destriping's
$0.74$--$0.80$)---at the price of inheriting destriping's own failure
modes when they occur (a plane extrapolated from clustered stable tiles
on the $16$-km cut-out, $337$; noisier strip medians at $3$--$5$ tiles,
$0.63$ against the fully-prior-constrained $0.54$; Supplementary
Table~S10). The choice between the two members is a statement about
which prior one trusts on the unidentifiable subspace; the paper's
claims rest on the identifiable part, where the two agree.

\subsection{From pairs to a network}
The three pairs of a triplet can be estimated jointly. 
We solve for $d_{AB}$ and $d_{BC}$ while imposing $d_{AC}=d_{AB}+d_{BC}$ exactly, assign one $\xi_p$ to each pair, and use the automatically selected priors. 
All unknowns are solved using a single Schur complement. 
The resulting photogrammetric network adjustment behaves as predicted by the closure analysis. 
It reduces the held-out noise floor by $4$--$10\%$ for the Ridgecrest triplet but produces no measurable reduction for Baltoro (Supplementary Table~S9). 
This difference occurs because $88\%$ of the pairwise error is associated with individual scenes and cancels in closure rather than averaging across pairs. 
The primary benefit of the joint solution is therefore consistency rather than a lower noise floor. 
The closure residual is identically zero, each time interval has one displacement field, and additional epochs, ground-control observations, or InSAR measurements can be introduced through further observation equations.

\section{Discussion}
\label{sec:disc}

\paragraph{Positioning against InSAR and offset tracking}
The proposed method complements InSAR. InSAR is most effective for small deformation over coherent terrain, whereas differential optical measurement addresses deformation with large spatial gradients and low radar coherence while providing an explicit error budget.
Single-pair accuracy is comparable to that of conventional offset tracking. Experiment E5 shows that destriping is a robust special case of the differential estimator, and the two approaches agree within their observed scatter for the Ridgecrest data.

The contribution is therefore not an unconditional improvement in single-pair accuracy. 
Instead, the differential model provides an estimation-based interpretation of destriping, a predictive error budget, pair-specific calibration of the matching uncertainty, and a unified interface for ground control, digital elevation models, multiple epochs, and InSAR.
It also provides design rules for the displacement-prior scale $\sigma_d$ and the viewing geometry. 
For imagery with a ground sampling distance of $10$\,m, the E4 simulation predicts a $2\sigma$ minimum detectable displacement of $0.38$\,m. 
The Ridgecrest experiment yields stable-area, MAD-based noise floors of $0.51$\,m for the destriping workflow and $0.48$--$0.54$\,m for the informed estimator under the automatic prior rule.

\paragraph{Honest boundaries.}
Three limits are stated plainly. The sub-meter envelope point is
verified in the across-look direction of one cross-view pair; a
same-view sub-meter pair remains the clean test, since the along-look
direction of a cross-view pair is bounded by resolution degradation
rather than by the method. The immunity and
control-free margins beyond the $10$\,m real floor rest on simulation
and structural argument---a rigorous real absolute baseline is not
reproducible on already-orthorectified ARD products, whose common
grid implicitly shares the datum. The blindness to mean along-track
displacement demonstrated in E5
is an information boundary shared by every control-free optical method,
including this one; a dense along-track signal requires ground control
or joint inversion with InSAR. A fourth limit is one of basis rather
than of principle: the revisit-offset basis used throughout is
per-line---bands perpendicular to the track---and does not contain the
detector-module offsets of the MSI, which are bands \emph{parallel} to
the track \citep{storey2016,stumpf2018}; on the full Ridgecrest tile
they are visible in the North component at $\pm0.3$--$0.5$\,m and are
the coherent residual noted in Section~\ref{sec:ridgecrest}. They are absorbed by adding a
second set of columns to $\bm B$ (bands parallel to the track, fitted
on stable tiles): in a first test this lowers the far-field North floor
from $0.44$ to $0.40$\,m and the North residual at the GNSS stations
from $0.27$ to $0.21$\,m, with the remaining structure varying along
the track within each band and calling for a module\,$\times$\,segment
basis. None of the comparisons in this paper is affected, since every
estimator was run on the same basis, but the absolute North floors
reported here are upper bounds for Sentinel-2.

\paragraph{Outlook: from single pairs to a monitoring instrument}
Three extensions follow directly from what is now measured. 
(i) \emph{Multi-year glacier series.} The Baltoro triplet is one season;
the same-orbit, same-season pairs available every autumn since 2016
give an interannual velocity series whose year-to-year change is
exactly the quantity that the closure law and the per-point $\sigma$
make testable rather than assumed---the observational input for the
slowdown-versus-thinning question raised by \citet{dehecq2019}, with
the caveat that velocity is a kinematic proxy: attributing its change
to climate requires a mass-balance model, which our product should
feed rather than replace. Vertical change from DEM differencing is the
natural companion product and is the same differential estimation with
elevation in place of the horizontal components. 
(ii) \emph{Extension to additional sensors} The envelope is written in GSD and matching noise,
so it transfers to any optical sensor; two further rungs of the
resolution ladder---a $2$\,m CBERS-04A triplet over a dune field and a
$0.45$\,m GeoEye-1 same-view null pair across the Sagaing fault
zone---are measured with the same pipeline and reported in
Supplementary Section~S3, so that $0.45$, $0.52$, $2$ and $10$\,m are
now covered and only the 16\,m wide-swath rung remains; the same-view
pairing rule and the fresh-snow rule are the acquisition constraints
to carry into tasking. 
(iii) \emph{Optical and InSAR fusion.} The per-point posterior $\sigma$ is the missing weight for joint optical--InSAR
inversion \citep{liu2026coal}, and the closure statistic is a scene
quality flag that a product chain such as ITS\_LIVE could carry with
every granule.

\paragraph{From first examples to an operational product}
Three steps separate what is demonstrated here from a product, and each
is a defined piece of work rather than an open question. \emph{(i)} The
per-line $\xi(t)$ extension has to be instantiated against a specific
sensor's attitude model; the closed-form plateau derived in
Section~\ref{sec:pushbroom} already fixes what that instantiation must
reproduce. \emph{(ii)} Cross-view pairs need DEM-coupled geometry; the
six-fold anisotropy measured here quantifies the error that coupling
must remove. \emph{(iii)} A per-sensor parameter table has to be
populated, for which the two-page calibration protocol of
Section~\ref{sec:method} is the recipe and the four rungs measured here
($0.45$, $0.52$, $2$, $10$\,m) are the first entries. The reward is a deformation product that
comes with its own error budget---for hazard response and
reconnaissance over control-denied areas, a capability the current
per-pair procedure does not offer.

\section{Conclusion}
\label{sec:conc}
This study formulates repeat-pass optical deformation measurement as a differential estimation problem free of surveyed ground control, in which nominal georeferencing defines the coordinate frame and stable-area constraints together with displacement priors resolve the datum.
The formulation absorbs inter-orbit biases, bounds along-track uncertainty through a displacement prior, represents pushbroom jitter using per-line revisit offsets, and quantifies leakage caused by view-angle and DEM errors. Three control-free real-data cases verify the predicted noise floor at both 10\,m and sub-meter resolutions. 
They also validate closure-based error decomposition, pair-specific matching-uncertainty calibration, the effect of cross-view geometry, and agreement with ITS\_LIVE glacier velocities. 
The experiments show that conventional destriping is a special case of the estimator and identify its limitations when stable terrain is sparse. 
They also identify the along-track mean of a dense deformation field as unobservable without additional information. 
Future work should extend the validation to multi-year glacier observations, additional resolution classes, and sensor-specific pushbroom models.

\section*{CRediT authorship contribution statement}
\textbf{Yueqiang Zhang}: Conceptualization, Methodology, Software,
Formal analysis, Writing -- original draft. \textbf{Chang Ma}:
Investigation, Data curation, Validation. \textbf{Shuixin Pan}:
Methodology, Supervision, Writing -- review \& editing, Funding
acquisition. \textbf{Haibo Liu}: Validation, Writing -- review \&
editing.

\section*{Declaration of competing interest}
The authors declare that they have no known competing financial
interests or personal relationships that could have appeared to
influence the work reported in this paper.

\section*{Data and code availability}
All imagery and reference datasets used in this study are publicly accessible. 
The processing scripts retrieve the following datasets through their respective public interfaces: Sentinel-2 L2A
Cloud-Optimized GeoTIFFs via the AWS Earth Search STAC API; Maxar Open
Data ARD (Kahramanmara\c{s} 2023, Myanmar 2025); CBERS-04A WPM via the
INPE Brazil Data Cube STAC; Copernicus GLO-30 DEM; ITS\_LIVE v2 velocity
pairs; RGI 6.0 outlines; USGS field displacement data for Ridgecrest.
The source code and experiment logs are available from the corresponding
author upon reasonable request.

\section*{Acknowledgements}
This work was supported in part by the National Natural Science Foundation of China under Grants 12372184 and 12002215, and in part by the Research Team Cultivation Program of Shenzhen University under Grant 2023JCT003.

\bibliographystyle{elsarticle-harv}
\bibliography{ref}

@article{ackermann1984,
  author  = {Friedrich Ackermann},
  title   = {Digital Image Correlation: Performance and Potential Application in Photogrammetry},
  journal = {The Photogrammetric Record},
  year    = {1984},
  volume  = {11},
  number  = {64},
  pages   = {429--439},
  doi     = {10.1111/j.1477-9730.1984.tb00505.x}
}

@article{vanpuymbroeck2000,
  author  = {Nad{\`e}ge {Van Puymbroeck} and R{\'e}mi Michel and Renaud Binet and Jean-Philippe Avouac and Jean Taboury},
  title   = {Measuring Earthquakes from Optical Satellite Images},
  journal = {Applied Optics},
  year    = {2000},
  volume  = {39},
  number  = {20},
  pages   = {3486--3494},
  doi     = {10.1364/AO.39.003486}
}

@article{michel2002,
  author  = {R{\'e}mi Michel and Jean-Philippe Avouac},
  title   = {Deformation Due to the 17 August 1999 Izmit, Turkey, Earthquake Measured from {SPOT} Images},
  journal = {Journal of Geophysical Research: Solid Earth},
  year    = {2002},
  volume  = {107},
  number  = {B4},
  pages   = {2062},
  doi     = {10.1029/2000JB000102}
}

@article{kaab2000,
  author  = {Andreas K{\"a}{\"a}b and Markus Vollmer},
  title   = {Surface Geometry, Thickness Changes and Flow Fields on Creeping Mountain Permafrost: Automatic Extraction by Digital Image Analysis},
  journal = {Permafrost and Periglacial Processes},
  year    = {2000},
  volume  = {11},
  number  = {4},
  pages   = {315--326},
  doi     = {10.1002/1099-1530(200012)11:4<315::AID-PPP365>3.0.CO;2-J}
}

@article{berthier2005,
  author  = {{\'E}tienne Berthier and H{\'e}l{\`e}ne Vadon and David Baratoux and Yves Arnaud and Christian Vincent and Kurt L. Feigl and Fr{\'e}d{\'e}rique R{\'e}my and Beno{\^\i}t Legr{\'e}sy},
  title   = {Surface Motion of Mountain Glaciers Derived from Satellite Optical Imagery},
  journal = {Remote Sensing of Environment},
  year    = {2005},
  volume  = {95},
  number  = {1},
  pages   = {14--28},
  doi     = {10.1016/j.rse.2004.11.005}
}

@article{debellagilo2011,
  author  = {Misganu Debella-Gilo and Andreas K{\"a}{\"a}b},
  title   = {Sub-Pixel Precision Image Matching for Measuring Surface Displacements on Mass Movements Using Normalized Cross-Correlation},
  journal = {Remote Sensing of Environment},
  year    = {2011},
  volume  = {115},
  number  = {1},
  pages   = {130--142},
  doi     = {10.1016/j.rse.2010.08.012}
}

@article{ayoub2009,
  author  = {Fran{\c{c}}ois Ayoub and S{\'e}bastien Leprince and Jean-Philippe Avouac},
  title   = {Co-Registration and Correlation of Aerial Photographs for Ground Deformation Measurements},
  journal = {ISPRS Journal of Photogrammetry and Remote Sensing},
  year    = {2009},
  volume  = {64},
  number  = {6},
  pages   = {551--560},
  doi     = {10.1016/j.isprsjprs.2009.03.005}
}

@article{stumpf2014,
  author  = {Andr{\'e} Stumpf and Jean-Philippe Malet and Pascal Allemand and Patrice Ulrich},
  title   = {Surface Reconstruction and Landslide Displacement Measurements with {Pl{\'e}iades} Satellite Images},
  journal = {ISPRS Journal of Photogrammetry and Remote Sensing},
  year    = {2014},
  volume  = {95},
  pages   = {1--12},
  doi     = {10.1016/j.isprsjprs.2014.05.008}
}

@article{dehecq2015,
  author  = {Amaury Dehecq and No{\"e}l Gourmelen and Emmanuel Trouv{\'e}},
  title   = {Deriving Large-Scale Glacier Velocities from a Complete Satellite Archive: Application to the Pamir--Karakoram--Himalaya},
  journal = {Remote Sensing of Environment},
  year    = {2015},
  volume  = {162},
  pages   = {55--66},
  doi     = {10.1016/j.rse.2015.01.031}
}

@article{millan2019,
  author  = {Romain Millan and J{\'e}r{\'e}mie Mouginot and Antoine Rabatel and Seongsu Jeong and Diego Cusicanqui and Anna Derkacheva and Mondher Chekki},
  title   = {Mapping Surface Flow Velocity of Glaciers at Regional Scale Using a Multiple Sensors Approach},
  journal = {Remote Sensing},
  year    = {2019},
  volume  = {11},
  number  = {21},
  pages   = {2498},
  doi     = {10.3390/rs11212498}
}

@article{lacroix2018,
  author  = {Pascal Lacroix and Gr{\'e}gory Bi{\`e}vre and Erwan Pathier and Ulrich Kniess and Denis Jongmans},
  title   = {Use of {Sentinel-2} Images for the Detection of Precursory Motions before Landslide Failures},
  journal = {Remote Sensing of Environment},
  year    = {2018},
  volume  = {215},
  pages   = {507--516},
  doi     = {10.1016/j.rse.2018.03.042}
}

@article{bontemps2018,
  author  = {No{\'e}lie Bontemps and Pascal Lacroix and Marie-Pierre Doin},
  title   = {Inversion of Deformation Fields Time-Series from Optical Images, and Application to the Long Term Kinematics of Slow-Moving Landslides in Peru},
  journal = {Remote Sensing of Environment},
  year    = {2018},
  volume  = {210},
  pages   = {144--158},
  doi     = {10.1016/j.rse.2018.02.023}
}

@article{aati2022,
  author  = {Saif Aati and Chris Milliner and Jean-Philippe Avouac},
  title   = {A New Approach for {2-D} and {3-D} Precise Measurements of Ground Deformation from Optimized Registration and Correlation of Optical Images and {ICA}-Based Filtering of Image Geometry Artifacts},
  journal = {Remote Sensing of Environment},
  year    = {2022},
  volume  = {277},
  pages   = {113038},
  doi     = {10.1016/j.rse.2022.113038}
}

@article{storey2016,
  author  = {James Storey and David P. Roy and Jeffrey Masek and Ferran Gascon and John Dwyer and Michael Choate},
  title   = {A Note on the Temporary Misregistration of {Landsat-8} Operational Land Imager ({OLI}) and {Sentinel-2} Multi Spectral Instrument ({MSI}) Imagery},
  journal = {Remote Sensing of Environment},
  year    = {2016},
  volume  = {186},
  pages   = {121--122},
  doi     = {10.1016/j.rse.2016.08.025}
}

@article{fraser1997,
  author  = {Clive S. Fraser},
  title   = {Digital Camera Self-Calibration},
  journal = {ISPRS Journal of Photogrammetry and Remote Sensing},
  year    = {1997},
  volume  = {52},
  number  = {4},
  pages   = {149--159},
  doi     = {10.1016/S0924-2716(97)00005-1}
}

@article{luhmann2016,
  author  = {Thomas Luhmann and Clive Fraser and Hans-Gerd Maas},
  title   = {Sensor Modelling and Camera Calibration for Close-Range Photogrammetry},
  journal = {ISPRS Journal of Photogrammetry and Remote Sensing},
  year    = {2016},
  volume  = {115},
  pages   = {37--46},
  doi     = {10.1016/j.isprsjprs.2015.10.006}
}

@article{scherler2011,
  author  = {Dirk Scherler and Bodo Bookhagen and Manfred R. Strecker},
  title   = {Spatially Variable Response of Himalayan Glaciers to Climate Change Affected by Debris Cover},
  journal = {Nature Geoscience},
  year    = {2011},
  volume  = {4},
  number  = {3},
  pages   = {156--159},
  doi     = {10.1038/ngeo1068}
}

@misc{pamidraft,
  author        = {Yueqiang Zhang and Liang Deng and Yi Zhang and Baoqiong Wang and Wenjun Chen and Shuixin Pan and Yulan Guo and Qifeng Yu},
  title         = {Differential {6-DOF} Pose Estimation with Provable First-Order Immunity to Camera Calibration Errors},
  year          = {2026},
  eprint        = {2608.04673},
  archiveprefix = {arXiv},
  primaryclass  = {cs.CV},
  doi           = {10.48550/arXiv.2608.04673},
  url           = {https://arxiv.org/abs/2608.04673},
  note          = {arXiv preprint}
}

@misc{lian2026,
  author        = {Meng Lian and Jian Wang and Shuixin Pan and Haibo Liu and Yueqiang Zhang and Yulan Guo},
  title         = {Beyond Control Points: Arcsecond Relative-Motion Estimation of Vision Measurement Platforms with Incomplete or Absent Control Fields},
  year          = {2026},
  eprint        = {2608.13918},
  archiveprefix = {arXiv},
  primaryclass  = {cs.CV},
  doi           = {10.48550/arXiv.2608.13918},
  url           = {https://arxiv.org/abs/2608.13918},
  note          = {arXiv preprint}
}

@article{meurant1992,
  author  = {G{\'e}rard Meurant},
  title   = {A Review on the Inverse of Symmetric Tridiagonal and Block Tridiagonal Matrices},
  journal = {SIAM Journal on Matrix Analysis and Applications},
  year    = {1992},
  volume  = {13},
  number  = {3},
  pages   = {707--728},
  doi     = {10.1137/0613045}
}

@article{leprince2007,
  author  = {S{\'e}bastien Leprince and Sylvain Barbot and Fran{\c{c}}ois Ayoub and Jean-Philippe Avouac},
  title   = {Automatic and Precise Orthorectification, Coregistration, and Subpixel Correlation of Satellite Images, Application to Ground Deformation Measurements},
  journal = {IEEE Transactions on Geoscience and Remote Sensing},
  year    = {2007},
  volume  = {45},
  number  = {6},
  pages   = {1529--1558},
  doi     = {10.1109/TGRS.2006.888937}
}

@incollection{avouac2014,
  author    = {Jean-Philippe Avouac and S{\'e}bastien Leprince},
  title     = {Geodetic Imaging Using Optical Systems},
  booktitle = {Treatise on Geophysics},
  editor    = {Gerald Schubert},
  edition   = {2},
  publisher = {Elsevier},
  address   = {Oxford},
  year      = {2015},
  pages     = {387--424},
  doi       = {10.1016/B978-0-444-53802-4.00067-1}
}

@article{scherler2008,
  author  = {Dirk Scherler and S{\'e}bastien Leprince and Manfred R. Strecker},
  title   = {Glacier-Surface Velocities in Alpine Terrain from Optical Satellite Imagery: Accuracy Improvement and Quality Assessment},
  journal = {Remote Sensing of Environment},
  year    = {2008},
  volume  = {112},
  number  = {10},
  pages   = {3806--3819},
  doi     = {10.1016/j.rse.2008.05.018}
}

@article{heid2012,
  author  = {Torborg Heid and Andreas K{\"a}{\"a}b},
  title   = {Evaluation of Existing Image Matching Methods for Deriving Glacier Surface Displacements Globally from Optical Satellite Imagery},
  journal = {Remote Sensing of Environment},
  year    = {2012},
  volume  = {118},
  pages   = {339--355},
  doi     = {10.1016/j.rse.2011.11.024}
}

@article{rosu2015,
  author  = {Ana-Maria Rosu and Marc Pierrot-Deseilligny and Arthur Delorme and Renaud Binet and Yann Klinger},
  title   = {Measurement of Ground Displacement from Optical Satellite Image Correlation Using the Free Open-Source Software {MicMac}},
  journal = {ISPRS Journal of Photogrammetry and Remote Sensing},
  year    = {2015},
  volume  = {100},
  pages   = {48--59},
  doi     = {10.1016/j.isprsjprs.2014.03.002}
}

@article{fahnestock2016,
  author  = {Mark Fahnestock and Ted Scambos and Twila Moon and Alex Gardner and Terry Haran and Marin Klinger},
  title   = {Rapid Large-Area Mapping of Ice Flow Using {Landsat 8}},
  journal = {Remote Sensing of Environment},
  year    = {2016},
  volume  = {185},
  pages   = {84--94},
  doi     = {10.1016/j.rse.2015.11.023}
}

@article{lei2021,
  author  = {Yang Lei and Alex Gardner and Piyush Agram},
  title   = {Autonomous Repeat Image Feature Tracking ({autoRIFT}) and Its Application for Tracking Ice Displacement},
  journal = {Remote Sensing},
  year    = {2021},
  volume  = {13},
  number  = {4},
  pages   = {749},
  doi     = {10.3390/rs13040749}
}

@article{kaab2016,
  author  = {Andreas K{\"a}{\"a}b and Solveig Winsvold and Bas Altena and Christopher Nuth and Thomas Nagler and Jan Wuite},
  title   = {Glacier Remote Sensing Using {Sentinel-2}. Part {I}: Radiometric and Geometric Performance, and Application to Ice Velocity},
  journal = {Remote Sensing},
  year    = {2016},
  volume  = {8},
  number  = {7},
  pages   = {598},
  doi     = {10.3390/rs8070598}
}

@article{altena2019,
  author  = {Bas Altena and Andreas K{\"a}{\"a}b},
  title   = {Weekly Glacier Flow Estimation from Dense Satellite Time Series Using Adapted Optical Flow Technology},
  journal = {Frontiers in Earth Science},
  year    = {2017},
  volume  = {5},
  pages   = {53},
  doi     = {10.3389/feart.2017.00053}
}

@article{stumpf2018,
  author  = {Andr{\'e} Stumpf and David Mich{\'e}a and Jean-Philippe Malet},
  title   = {Improved Co-Registration of {Sentinel-2} and {Landsat-8} Imagery for Earth Surface Motion Measurements},
  journal = {Remote Sensing},
  year    = {2018},
  volume  = {10},
  number  = {2},
  pages   = {160},
  doi     = {10.3390/rs10020160}
}

@article{dehecq2019,
  author  = {Amaury Dehecq and No{\"e}l Gourmelen and Alex S. Gardner and Fanny Brun and Daniel Goldberg and Peter W. Nienow and {\'E}tienne Berthier and Christian Vincent and Patrick Wagnon and Emmanuel Trouv{\'e}},
  title   = {Twenty-First Century Glacier Slowdown Driven by Mass Loss in High Mountain Asia},
  journal = {Nature Geoscience},
  year    = {2019},
  volume  = {12},
  number  = {1},
  pages   = {22--27},
  doi     = {10.1038/s41561-018-0271-9}
}

@article{millan2022,
  author  = {Romain Millan and J{\'e}r{\'e}mie Mouginot and Antoine Rabatel and Mathieu Morlighem},
  title   = {Ice Velocity and Thickness of the World's Glaciers},
  journal = {Nature Geoscience},
  year    = {2022},
  volume  = {15},
  number  = {2},
  pages   = {124--129},
  doi     = {10.1038/s41561-021-00885-z}
}

@article{hugonnet2021,
  author  = {Romain Hugonnet and Robert McNabb and {\'E}tienne Berthier and Brian Menounos and Christopher Nuth and Luc Girod and Daniel Farinotti and Matthias Huss and Ines Dussaillant and Fanny Brun and Andreas K{\"a}{\"a}b},
  title   = {Accelerated Global Glacier Mass Loss in the Early Twenty-First Century},
  journal = {Nature},
  year    = {2021},
  volume  = {592},
  number  = {7856},
  pages   = {726--731},
  doi     = {10.1038/s41586-021-03436-z}
}

@article{milliner2015,
  author  = {Christopher W. D. Milliner and James F. Dolan and James Hollingsworth and S{\'e}bastien Leprince and Fran{\c{c}}ois Ayoub and Charles G. Sammis},
  title   = {Quantifying Near-Field and Off-Fault Deformation Patterns of the 1992 {$M_{\mathrm{w}}$ 7.3} Landers Earthquake},
  journal = {Geochemistry, Geophysics, Geosystems},
  year    = {2015},
  volume  = {16},
  number  = {5},
  pages   = {1577--1598},
  doi     = {10.1002/2014GC005693}
}

@article{milliner2020,
  author  = {Chris Milliner and Andrea Donnellan},
  title   = {Using Daily Observations from Planet Labs Satellite Imagery to Separate the Surface Deformation between the 4 July {$M_{\mathrm{w}}$ 6.4} Foreshock and 5 July {$M_{\mathrm{w}}$ 7.1} Mainshock during the 2019 Ridgecrest Earthquake Sequence},
  journal = {Seismological Research Letters},
  year    = {2020},
  volume  = {91},
  number  = {4},
  pages   = {1986--1997},
  doi     = {10.1785/0220190271}
}

@article{blewitt2018,
  author  = {Geoffrey Blewitt and William C. Hammond and Corn{\'e} Kreemer},
  title   = {Harnessing the {GPS} Data Explosion for Interdisciplinary Science},
  journal = {Eos},
  year    = {2018},
  volume  = {99},
  doi     = {10.1029/2018EO104623}
}

@article{barnhart2019,
  author  = {William D. Barnhart and Gavin P. Hayes and Ryan D. Gold},
  title   = {The July 2019 Ridgecrest, California, Earthquake Sequence: Kinematics of Slip and Stressing in Cross-Fault Ruptures},
  journal = {Geophysical Research Letters},
  year    = {2019},
  volume  = {46},
  number  = {21},
  pages   = {11859--11867},
  doi     = {10.1029/2019GL084741}
}

@article{barbot2023,
  author  = {Sylvain Barbot and Heng Luo and Teng Wang and Yariv Hamiel and Oksana Piatibratova and Muhammad Tahir Javed and Carla Braitenberg and G{\"o}khan G{\"u}rb{\"u}z},
  title   = {Slip Distribution of the February 6, 2023 {$M_{\mathrm{w}}$ 7.8} and {$M_{\mathrm{w}}$ 7.6}, Kahramanmara{\c{s}}, Turkey Earthquake Sequence in the East Anatolian Fault Zone},
  journal = {Seismica},
  year    = {2023},
  volume  = {2},
  number  = {3},
  doi     = {10.26443/seismica.v2i3.502}
}

@article{lacroix2019,
  author  = {Pascal Lacroix and Alexander L. Handwerger and Gr{\'e}gory Bi{\`e}vre},
  title   = {Life and Death of Slow-Moving Landslides},
  journal = {Nature Reviews Earth \& Environment},
  year    = {2020},
  volume  = {1},
  number  = {8},
  pages   = {404--419},
  doi     = {10.1038/s43017-020-0072-8}
}

@article{insar_review,
  author  = {Michele Crosetto and Oriol Monserrat and Mar{\'i}a Cuevas-Gonz{\'a}lez and N{\'u}ria Devanth{\'e}ry and Bruno Crippa},
  title   = {Persistent Scatterer Interferometry: A Review},
  journal = {ISPRS Journal of Photogrammetry and Remote Sensing},
  year    = {2016},
  volume  = {115},
  pages   = {78--89},
  doi     = {10.1016/j.isprsjprs.2015.10.011}
}

@article{liu2026coal,
  author  = {Zihan Liu and Junhuan Peng and Huiwei Su and Xiang Li and Chenyu Wang and Yun Peng},
  title   = {{InSAR} Monitoring of Large Gradient Deformation in Coalfield and Phase Unwrapping Research},
  journal = {Geodesy and Geodynamics},
  year    = {2026},
  volume  = {17},
  number  = {1},
  pages   = {130--144},
  doi     = {10.1016/j.geog.2025.06.001}
}

@article{chen2026edge,
  author  = {Liquan Chen and Jing Hu and Zhong Lu and Xuhao Li and Chaoying Zhao and Guangrong Li and Jifei Wang},
  title   = {Edge Detection-Based Network Optimization Method for High-Gradient {InSAR} Deformation Phase Unwrapping},
  journal = {International Journal of Remote Sensing},
  year    = {2026},
  pages   = {1--14},
  note    = {Advance online publication},
  doi     = {10.1080/01431161.2026.2707263}
}

@article{rouetleduc2025unwrap,
  author  = {Bertrand Rouet-Leduc and Claudia Hulbert},
  title   = {Phase Unwrapping in Seconds: A Spectral {ADMM} Algorithm for Large-Scale {InSAR}},
  journal = {Remote Sensing},
  year    = {2026},
  volume  = {18},
  number  = {11},
  pages   = {1801},
  doi     = {10.3390/rs18111801}
}

@article{antoine2025topo,
  author  = {Sol{\`e}ne L. Antoine and Zhen Liu},
  title   = {Impact of Optical Imagery and Topography Data Resolution on the Measurement of Surface Fault Displacement Using Sub-Pixel Image Correlation},
  journal = {Earth and Space Science},
  year    = {2025},
  volume  = {12},
  number  = {4},
  pages   = {e2024EA003660},
  doi     = {10.1029/2024EA003660}
}

@article{montagnon2024deep,
  author  = {Tristan Montagnon and Sophie Giffard-Roisin and Mauro {Dalla Mura} and Mathilde Marchandon and Erwan Pathier and James Hollingsworth},
  title   = {Sub-Pixel Displacement Estimation with Deep Learning: Application to Optical Satellite Images Containing Sharp Displacements},
  journal = {Journal of Geophysical Research: Machine Learning and Computation},
  year    = {2024},
  volume  = {1},
  number  = {4},
  pages   = {e2024JH000174},
  doi     = {10.1029/2024JH000174}
}

@article{han2023menyuan,
  author  = {Nana Han and Guohong Zhang and Xinjian Shan and Yingfeng Zhang and Eric Hetland and Chunyan Qu and Wenyu Gong and Guangtong Sun and Chenglong Li and Xiaoran Fan and Chuanchao Huang},
  title   = {Coseismic Surface Horizontal Deformation of the 2022 {$M_{\mathrm{w}}$ 6.6} Menyuan, Qinghai, China, Earthquake from Optical Pixel Correlation of {GF-7} Stereo Satellite Images},
  journal = {Seismological Research Letters},
  year    = {2023},
  volume  = {94},
  number  = {4},
  pages   = {1747--1760},
  doi     = {10.1785/0220220332}
}

@article{chang2024stacking,
  author  = {Fengnian Chang and Shaochun Dong and Hongwei Yin and Xiao Ye and Wei Zhang and Honghu Zhu},
  title   = {Temporal Stacking of Sub-Pixel Offset Tracking for Monitoring Slow-Moving Landslides in Vegetated Terrain},
  journal = {Landslides},
  year    = {2024},
  volume  = {21},
  number  = {6},
  pages   = {1255--1271},
  doi     = {10.1007/s10346-024-02227-7}
}

@article{zhou2025corner,
  author  = {Dingyi Zhou and Zhifang Zhao and Fei Zhao},
  title   = {Improved Pixel Offset Tracking Method Based on Corner Point Variation in Large-Gradient Landslide Deformation Monitoring},
  journal = {Remote Sensing},
  year    = {2025},
  volume  = {17},
  number  = {19},
  pages   = {3292},
  doi     = {10.3390/rs17193292}
}

@article{zheng2023glaft,
  author  = {Whyjay Zheng and Shashank Bhushan and Maximillian {Van Wyk De Vries} and William Kochtitzky and David Shean and Luke Copland and Christine Dow and Renette Jones-Ivey and Fernando P{\'e}rez},
  title   = {{GLAcier Feature Tracking} Testkit ({GLAFT}): A Statistically and Physically Based Framework for Evaluating Glacier Velocity Products Derived from Optical Satellite Image Feature Tracking},
  journal = {The Cryosphere},
  year    = {2023},
  volume  = {17},
  number  = {9},
  pages   = {4063--4078},
  doi     = {10.5194/tc-17-4063-2023}
}

@article{zhang2024karakoram,
  author  = {Zhijie Zhang and Zahoor Ahmad and Shengqin Xiong and Wanchang Zhang},
  title   = {Glacier Velocity and Surge Detection in the Karakoram Region, Pakistan: Using Remotely Sensed Data with Cross-Correlation Feature Tracking},
  journal = {International Journal of Digital Earth},
  year    = {2024},
  volume  = {17},
  number  = {1},
  pages   = {2441928},
  doi     = {10.1080/17538947.2024.2441928}
}

@article{wang2025haiyang,
  author  = {Yanli Wang and Ronghao Zhang and Yizhang Xu and Xiangyu Zhang and Rongfan Dai and Shuying Jin},
  title   = {Jitter Error Correction for the {HaiYang-3A} Satellite Based on Multi-Source Attitude Fusion},
  journal = {Remote Sensing},
  year    = {2025},
  volume  = {17},
  number  = {9},
  pages   = {1489},
  doi     = {10.3390/rs17091489}
}

@article{xia2024jitter,
  author  = {Haoran Xia and Xinming Tang and Fan Mo and Junfeng Xie and Xiang Li},
  title   = {Geographically-Informed Modeling and Analysis of Platform Attitude Jitter in {GF-7} Sub-Meter Stereo Mapping Satellite},
  journal = {ISPRS International Journal of Geo-Information},
  year    = {2024},
  volume  = {13},
  number  = {11},
  pages   = {413},
  doi     = {10.3390/ijgi13110413}
}

@article{li2025dynamic,
  author  = {Da Li and Yonghua Jiang and Jingyin Wang and Shaodong Wei and Guo Zhang and Huilong Wang},
  title   = {A Generic Compensation Method for Dynamic Systematic Errors in the Geolocation of Linear Pushbroom Satellite Imagery},
  journal = {Planetary and Space Science},
  year    = {2025},
  volume  = {267},
  pages   = {106139},
  doi     = {10.1016/j.pss.2025.106139}
}
\end{document}


\begin{frontmatter}
\title{Supplementary material for ``Spaceborne differential
photogrammetry for control-free measurement of large-gradient
deformation with structural immunity and a predictable accuracy
envelope''}
\tnotetext[tnote1]{Submitted to \textit{ISPRS Journal of Photogrammetry and Remote Sensing}.}
\author[szu]{Yueqiang Zhang}\author[szu]{Chang Ma}\author[szu]{Shuixin Pan\corref{cor1}}
\ead{shuixinpan@szu.edu.cn}\author[hnu]{Haibo Liu}
\cortext[cor1]{Corresponding author.}
\affiliation[szu]{organization={State Key Laboratory of Radio Frequency Heterogeneous Integration; Key Laboratory of Optoelectronic Devices and Systems of Ministry of Education and Guangdong Province; Shenzhen Key Laboratory of Intelligent Optical Measurement and Detection; College of Physics and Optoelectronic Engineering, Shenzhen University}, city={Shenzhen}, postcode={518060}, country={China}}
\affiliation[hnu]{organization={School of Artificial Intelligence and Robotics, Hunan University}, city={Changsha},
postcode={410082}, country={China}}
\end{frontmatter}

\renewcommand{\thesection}{S\arabic{section}}
\renewcommand{\thefigure}{\thesection}
\renewcommand{\thetable}{\thesection}

\section{Uncertainty calibration on four scene pairs}
\label{sec:s1}
We repeated the two-band calibration used for the Ridgecrest analysis in the main text using four scene pairs (Ridgecrest L1C and L2A, Baltoro 2024 and 2025). The relation is monotone everywhere; fitted power-law exponents range from $0.24$ to $0.65$, so the calibration is per pair.

\begin{figure}[H]
\centering
\includegraphics[width=0.55\textwidth]{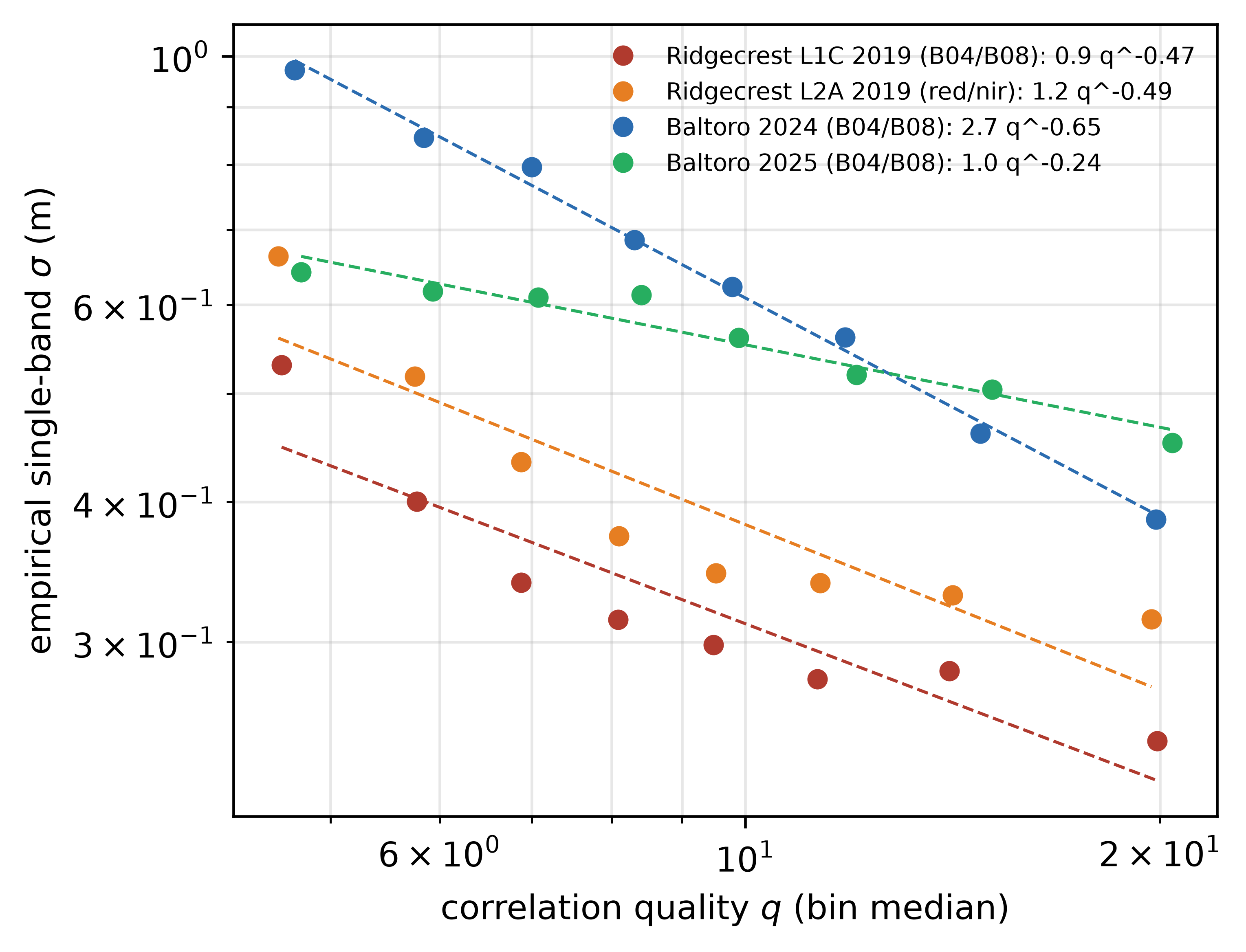}
\caption{The same calibration on four scene pairs. The relation is monotone everywhere; its coefficients are scene-specific (exponents $0.24$--$0.65$), so $\sigma(q)$ is calibrated per pair from the two-band redundancy rather than transferred as a constant.}
\label{fig:s1}
\end{figure}

\section{Why the intraseasonal analysis is not reported}
\label{sec:s2}
We also investigated intraseasonal dynamics using all cloud-free image pairs with temporal intervals of 5--15 days from seven autumn seasons between 2019 and 2025. 
The dataset comprises 33 pairs, including cross-platform observations from Sentinel-2A and Sentinel-2B. 
Trunk velocities estimated from pairs with intervals of 5, 10, and 15 days over comparable periods differ by factors of up to two. 
Some pairs also produce negative projections in the along-flow direction. 
These inconsistencies indicate residual coregistration offsets of $0.5$--$1$\,m that leak into the glacier-displacement estimates rather than actual glacier dynamics.

The study area is $57\%$ glacierized and contains only $54$ usable bedrock tiles. 
These observations are insufficient to constrain the trend and strip corrections below the displacement expected over a ten-day interval. 
Consequently, the formal uncertainty of approximately $2$\,m\,yr$^{-1}$ for the median trunk velocity understates the actual error by approximately one order of magnitude when the correction is weakly determined. 
The closure statistic identifies this failure mode.
The 2025 same-platform triplet yields closure ratios of $0.38$ and $0.60$, whereas the cross-platform pairs do not satisfy the closure test. 
Week-scale analysis of a debris-covered glacier at $10$\,m resolution therefore requires more extensive bedrock support or a same-platform acquisition protocol with a fixed temporal interval. 
We draw no conclusion about intraseasonal deceleration from these data.

\section{Additional sensors at 2 and 0.45\,m resolution}
\label{sec:s3}
We applied the same processing pipeline to two additional optical sensors, extending the evaluated ground sampling distances to $2$ and $0.45$\,m. 
At $2$\,m resolution, a CBERS-04A WPM triplet over the Len\c{c}\'ois Maranhenses dune field produces a noise floor of $0.13$--$0.20$\,m, equivalent to $0.06$--$0.10$ pixels, over the vegetated hinterland. 
The two closure ratios are $0.68$ and $0.37$.
The estimated dune-migration azimuth is $220^\circ$, compared with the expected trade-wind direction of $225^\circ$. 
A Sentinel-2 pair covering the same image window gives an azimuth of $209^\circ$.

At $0.45$\,m resolution, we analyze a GeoEye-1 pair acquired by the same sensor and from the same viewing direction across the Sagaing fault zone. 
Both images were acquired before the 2025 Myanmar earthquake and are separated by 11 days. 
The null pair has directional noise floors of $0.30$ and $0.36$ pixels, equivalent to $0.14$ and $0.16$\,m, and an anisotropy ratio of $1.21$. 
This result provides a second validation of the recommendation to use imagery acquired from similar viewing directions. 
The experiments now cover ground sampling distances of $0.45$, $0.52$, $2$, and $10$\,m. 
Only the $16$\,m wide-swath class remains unevaluated. 
Similar viewing directions and the avoidance of snowfall between acquisitions should therefore be included in future tasking criteria.

\begin{figure}[H]
\centering
\includegraphics[width=\textwidth]{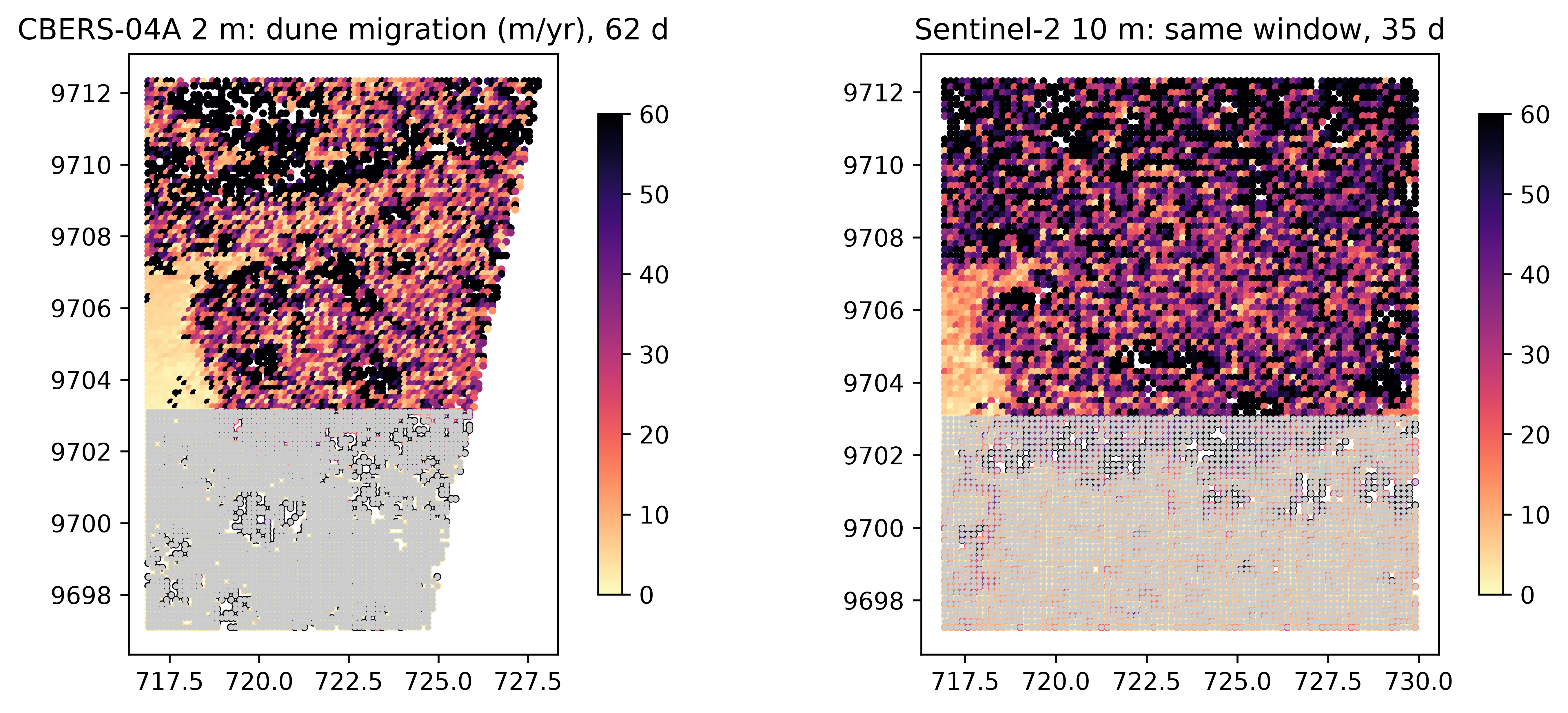}
\caption{CBERS-04A 2\,m (left) and Sentinel-2 10\,m (right) dune migration over the same window; gray = vegetated stable area.}
\label{fig:s3}
\end{figure}

\section{Simulation protocol and complete results}
\label{sec:s4}
The Monte-Carlo envelope of the main text (E1--E5) uses a first-order
single-component pushbroom model: $T=40$ trials with $2000$ bootstrap
resamples per statistic; $20\times20$ ground grid; $H=500$\,km;
baseline orbit error $5$\,cm, attitude $2''$, drift $1''$/epoch, line
jitter $0.2''$, digital elevation model (DEM) error $10$\,m, matching noise $0.1$\,px. 
Injection levels, mismatch sweeps, and full result tables can be reproduced
using the experiment scripts described in Section~\ref{sec:s5}; the numbers
quoted in the main text are those produced by these scripts.

\section{Reproducibility}
\label{sec:s5}
\begin{sloppypar}
Every real-data figure in the main text can be regenerated using a
corresponding script. These scripts are
\path{run_ridgecrest.py}, \path{run_closure.py},
\path{run_deepdig23.py}, \path{run_maxar*.py},
\path{run_glacier*.py}, \path{run_itslive_closure.py},
\path{run_calib_universal.py}, \path{run_headtohead.py},
\path{run_cbers_dunes.py}, and \path{run_maxar_myanmar_null.py}.
Imagery is streamed by HTTP range requests, so no dataset needs to be
downloaded in full. The experiment logs record every prespecified
criterion and every correction made during the study, including the
withdrawal of an earlier power-law summary of the uncertainty
calibration and of an intraseasonal glacier claim.
The source code and experiment logs are available from the corresponding
author upon reasonable request.
\end{sloppypar}

\section{Sensitivity to the prior scale}
\label{sec:s6}
The unknown-tile displacement prior $\sigma_d$ was varied from $3$ to
$100$\,m, with a strip prior of $0.2$--$0.5$\,m and stable tiles fixed
at $0.05$\,m. Table~\ref{tab:s6} reports the injected-truth root-mean-square error (RMSE) on the
real Ridgecrest background with robust weighting disabled.
\begin{table}[H]
\centering
\small
\caption{Injected-truth RMSE on the real Ridgecrest background for different displacement-prior scales. All values are in meters.}
\label{tab:s6}
\begin{tabular}{lccccc}\toprule
field / stable & $3$ & $10$ & $20$ & $50$ & $100$\,m \\\midrule
local patch, 100\% & 0.59 & 0.48 & 0.46 & 0.46 & 0.46 \\
local patch, 5\% & 0.76 & 0.69 & 0.67 & 0.67 & 0.67 \\
smooth field, 100\% & 1.20 & 1.19 & 1.18 & 1.12 & 0.99 \\
smooth field, 5\% & 1.20 & 1.19 & 1.17 & 1.13 & 1.03 \\
strip band, 100\% & 1.40 & 1.19 & 0.97 & 0.82 & 0.79 \\
strip band, 5\% & 1.45 & 1.32 & 1.09 & 0.82 & 0.76 \\\bottomrule
\end{tabular}
\end{table}
The Baltoro pair exhibits the opposite trend. When only $5$--$13$ stable tiles are available, the median trunk velocity is $82$--$85$\,m\,yr$^{-1}$ at $\sigma_d=4$\,m because part of the glacier motion is absorbed by the trend plane. 
The velocity increases to $85$--$89$\,m\,yr$^{-1}$ when $\sigma_d$ is between $20$ and $30$\,m, and to $96$--$135$\,m\,yr$^{-1}$ when $\sigma_d=100$\,m because the plane is insufficiently regularized. 
Setting the prior scale to approximately ten times the expected displacement provides a compromise between deformation attenuation and insufficient regularization. Both limiting behaviors are evident in the measured data.

\section{Natural image subsets from a second season and a coseismic scene}
\label{sec:s7}
\begin{table}[H]
\centering
\small
\caption{Natural cut-outs, no thinning; automatic priors. Baltoro 2025 AB: trunk median (m\,yr$^{-1}$) / shape root-mean-square (RMS) difference (m) against the full-scene 51-tile masked solution. Ridgecrest B04: cross-fault step at the maximum-slip point (m) / shape RMS difference (m) against the full-scene 5420-tile masked solution.}
\label{tab:s7}
\begin{adjustbox}{max width=\textwidth}
\begin{tabular}{llcccc}\toprule
Scene & window (stable tiles) & plane+strip & constant+strip & informed (auto) & reference \\\midrule
Baltoro 2025 & 8\,km (9) & 553 / 10.9 & 90 / 1.03 & 72 / 0.98 & 83 \\
 & 12\,km (11) & 725 / 18.3 & 100 / 1.02 & 75 / 0.90 & 79 \\
 & 16\,km (24) & 122 / 3.39 & 78 / 0.98 & 68 / 1.02 & 77 \\
 & 16\,km (4) & 776 / 14.1 & 95 / 0.90 & 72 / 0.86 & 69 \\
 & 8\,km (0) & -- / 1.22 & -- / 0.92 & -- / 0.43 & \\
 & 8\,km (0) & -- / 1.07 & -- / 1.01 & -- / 0.34 & \\
 & 12\,km (0) & -- / 1.21 & -- / 1.16 & -- / 0.83 & \\
 & 12\,km (0) & -- / 0.94 & -- / 0.85 & -- / 0.45 & \\
 & 16\,km (0) & -- / 1.15 & -- / 1.15 & -- / 0.60 & \\
Ridgecrest & 8\,km at rupture (0) & 3.15 / 0.92 & 3.31 / 0.63 & 3.75 / 0.23 & 3.75 \\
 & 12\,km at rupture (1) & 3.39 / 0.83 & 3.84 / 0.42 & 3.75 / 0.31 & 3.75 \\
 & 16\,km at rupture (61) & 3.89 / 0.84 & 3.88 / 0.29 & 3.79 / 0.24 & 3.75 \\
 & 24\,km at rupture (561) & 3.88 / 0.58 & 3.88 / 0.37 & 3.82 / 1.00 & 3.75 \\
 & 8\,km, $-6$\,km along strike (0) & 0.02 / 0.60 & 0.34 / 0.60 & 0.14 / 0.32 & 0.38 \\
 & 8\,km, $+6$\,km along strike (0) & 0.79 / 0.67 & 0.56 / 0.51 & 1.10 / 0.36 & 1.31 \\
 & 12\,km, $-6$\,km (0) & 0.06 / 0.57 & 0.28 / 0.56 & 0.17 / 0.38 & 0.38 \\
 & 12\,km, $+6$\,km (7) & 0.89 / 3.33 & 1.40 / 0.32 & 1.16 / 0.36 & 1.31 \\
 & 16\,km, $-6$\,km (18) & 0.15 / 0.81 & 0.37 / 0.22 & 0.21 / 0.35 & 0.38 \\
 & 16\,km, $+6$\,km (172) & 1.29 / 1.01 & 1.40 / 0.33 & 1.31 / 1.08 & 1.31 \\
 & 24\,km, $-6$\,km (572) & 0.22 / 0.34 & 0.25 / 0.40 & 0.24 / 0.72 & 0.38 \\
 & 24\,km, $+6$\,km (1202) & 1.32 / 0.31 & 1.14 / 0.44 & 0.81 / 0.58 & 1.31 \\\bottomrule
\end{tabular}
\end{adjustbox}
\end{table}

\section{Comparison of manually and automatically selected priors}
\label{sec:s8}
\begin{table}[H]
\centering
\rotatebox{90}{%
\begin{minipage}{0.94\textheight}
\centering
\small
\caption{Hand-set versus automatic priors on every prior-constrained experiment. Floors are stable-area Gaussian-scaled median absolute deviations ($\mathrm{MAD}_\sigma$; m); trunk speeds in m\,yr$^{-1}$; RMSE/RMS in m. ``rule'' applies the robust switch as stated; ``L2'' forces L2 where the switch chose Huber.}
\label{tab:s8}
\begin{adjustbox}{max width=\linewidth}
\begin{tabular}{llccc}\toprule
Experiment & quantity & destriping & informed, hand-set & informed, automatic \\\midrule
Ridgecrest B04 & far-field floor E/N & 0.51 / 0.56 & 0.41 / 0.41 ($\sigma_d$=2) & 0.48 / 0.47 (pinned half) \\
 & cross-fault step & $+3.16$ & $+2.50$ & $+3.14$ \\
Thinning, Ridgecrest B08 & floor at 50\%$\to$2\% stable & 0.51$\to$1.31 & 0.51 flat & 0.50$\to$0.57 \\
Thinning, Baltoro & trunk at 5 stable tiles & 1251 & 87 & 87 (L2) / 70 (rule: Huber) \\
 & trunk, no stable tile & 59 (full-field) & 86 & 87 (L2) / 66 (rule) \\
Injected truth & patch, 100/20/5\% & 0.65/0.67/0.69 & 0.49/0.66/0.71 & 0.54/0.69/0.74 \\
 & smooth field, 37/20 tiles & 0.79/2.47 & 1.19/1.19 & 1.19/1.19 \\
 & strip-aligned band, 100/5\% & 0.80/0.74 & 1.16/1.30 & 1.33/1.39 \\
Natural cut-outs & corner-bedrock trunk (ref.\ 83/82/80) & 647/778/337 & 82/88/95 & 88/78/86 \\
 & bedrock-free shape RMS (6 windows) & 1.22--1.92 & 0.84--1.65 & 0.83--1.66 \\\bottomrule
\end{tabular}
\end{adjustbox}
\end{minipage}%
}
\end{table}

\section{Triplet adjustment with a closure constraint}
The held-out automatic-prior statistics reported below are not directly comparable to the all-stable-tile closure statistics in the main text.
The former evaluate held-out halves of the stable set, whereas the latter use the full stable set and the baseline estimator.
\label{sec:s9}
\begin{table}[H]
\centering
\small
\caption{Per-pair and joint solutions with automatic priors. Floors and
closure values are Gaussian-scaled median absolute deviations
($\mathrm{MAD}_\sigma$) of the fault-parallel component evaluated on
held-out stable tiles. The signal is the Ridgecrest cross-fault step or
the Baltoro trunk median.}
\label{tab:s9}
\begin{adjustbox}{max width=\textwidth}
\begin{tabular}{llcccc}\toprule
Scene & setting & floor AB / BC / AC (m) & closure MAD$\sigma$ & signal AB / BC / AC \\\midrule
Ridgecrest, 3/8/13 July 2019 & per pair & 0.449 / 0.263 / 0.435 & 0.306 & 3.82 / $-0.01$ / 3.57 \\
 & joint & 0.407 / 0.254 / 0.408 & 0 & 3.79 / $-0.02$ / 3.58 \\
Baltoro 2024 & per pair & 1.176 / 1.704 / 2.291 & 0.762 & 87 / 110 / 90 \\
 & joint & 1.152 / 1.754 / 2.481 & 0 & 90 / 114 / 98 \\\bottomrule
\end{tabular}
\end{adjustbox}
\end{table}

\section{Strict hybrid estimator}
\label{sec:s10}
\begin{table}[H]
\centering
\rotatebox{90}{%
\begin{minipage}{0.94\textheight}
\centering
\small
\caption{Strict hybrid (masked destriping on supported strips and, when the stable tiles span the window, the plane; informed prior only on unsupported strips) against masked destriping and the fully prior-constrained estimator with automatic priors. Same experiments and metrics as the main text.}
\label{tab:s10}
\begin{adjustbox}{max width=\linewidth}
\begin{tabular}{llccc}\toprule
Experiment & quantity & destriping & prior-constrained (auto) & strict hybrid \\\midrule
Thinning, Ridgecrest B08 & floor at 50/20/10/5/2\% & 0.51/0.65/0.58/0.63/1.31 & 0.50/0.51/0.53/0.54/0.57 & 0.51/0.65/0.58/0.63/1.19 \\
Thinning, Baltoro & trunk at 27/13/5 tiles & 150/181/1251 & 98/82/87 & 113/100/92 \\
Injected truth & patch, 100/20/5\% & 0.65/0.67/0.69 & 0.54/0.69/0.74 & 0.46/0.65/0.69 \\
 & smooth field, 37/20 tiles & 0.79/2.47 & 1.19/1.19 & 1.42/1.36 \\
 & strip-aligned band, 100/20/5\% & 0.80/0.78/0.74 & 1.33/1.37/1.39 & 0.73/0.77/0.74 \\
Natural cut-outs, Baltoro & corner-bedrock trunk 8/12/16\,km (ref.\ 83/82/80) & 647/778/337 (plane) & 88/78/86 & 101/109/337 \\
 & bedrock-free shape RMS (6 windows) & 1.22--1.92 & 0.83--1.66 & 0.83--1.66 (identical: no supported strip) \\\bottomrule
\end{tabular}
\end{adjustbox}
\end{minipage}%
}
\end{table}

\section{Zero-deformation control for CBERS-04A imagery}
\label{sec:s11}

The 2\,m rung of the resolution ladder in the main text rests on a dune
field (path/row 204/116), where no genuinely stable surface exists and
the stable set had to be taken as the southernmost 40\% of the window---a
geographic proxy chosen partly because it moved least, which risks a
selection bias in its favor. We therefore processed the adjacent frame,
204/117, on the same two dates (28 October and 28 November 2025, 31 days), which
covers inland savanna and cropland rather than dunes. 
It contains abundant stable terrain and no expected deformation, providing a suitable control scene.

Of the frame, $78\%$ is cloud-free in both epochs; we take a
$22\times22$\,km window that is $98.6\%$ clear in both. Matching on the
same $64$-px chips and $32$-px step returns valid offsets at $33.4\%$ of
the nodes---tropical vegetation decorrelates over 31 days, and the
correlation is carried by roads, field boundaries and bare ground. The
stable set is split at random, half to fit the corrections and half held
out for evaluation ($19{,}606$ nodes).

\begin{figure}[H]\centering
\includegraphics[width=\linewidth]{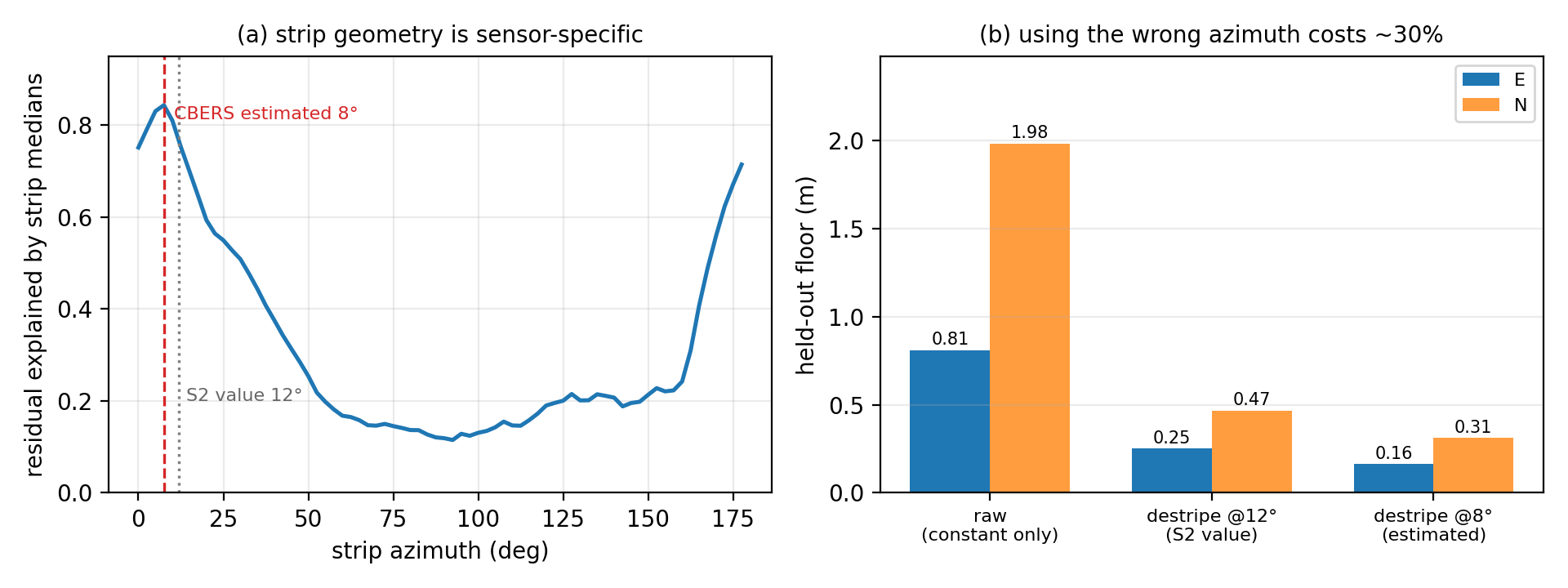}
\caption{CBERS-04A 204/117. (a) Fraction of the detrended residual
explained by strip medians against assumed strip azimuth; the maximum is
at $7.5^\circ$, not at the Sentinel-2 value of $12^\circ$. (b) Held-out
floors: removing a constant only, destriping at the Sentinel-2 azimuth,
and destriping at the estimated azimuth.}
\label{fig:s11}
\end{figure}

Two results follow (Table~\ref{tab:s11}, Fig.~\ref{fig:s11}).

First, the correction fitted over the full field and the masked
correction fitted over stable terrain agree within $0.001$ pixels, with
noise floors of $0.179$ and $0.180$ pixels, respectively. This result
reproduces the special-case relation derived in the main text using a
different sensor, continent, and land-cover type. When a scene contains
no measurable deformation and sufficient stable terrain is available,
the two correction procedures are equivalent. The present experiment
also quantifies these conditions. Among the $60$ along-track strips
crossing the image window, only $2$, or $3\%$, contain fewer than three
valid nodes, while the median strip contains $145$ valid nodes.

Second, the strip geometry is sensor-specific and must be estimated from
the data. Scanning the azimuth and asking how much of the detrended
residual is explained by strip medians gives a maximum at
$7.5^\circ$ ($84\%$ explained), not at the $12^\circ$ that applies to the
Sentinel-2 detector modules. Destriping at the Sentinel-2 azimuth still
works---and still agrees with the masked correction, since both are given
the same model---but it leaves a floor of $0.250/0.467$\,m against
$0.162/0.311$\,m at the estimated azimuth, a penalty of about $30\%$. 
In the implementation, the azimuth is therefore estimated per scene rather than configured.

Taken with the dune-field pair, the 2\,m floor is $0.065$--$0.085$\,px on
dunes and $0.081$--$0.156$\,px here: a factor of about two between two
scenes at the same resolution and sensor, both inside the
$0.04$--$0.27$\,px band predicted by the envelope. The spread between
scenes is the reason the envelope is a band and not a line.

\begin{table}[H]\centering\footnotesize
\caption{CBERS-04A 204/117, 28 October to 28 November 2025, $2$\,m, held-out
stable nodes. Floors in m.}
\label{tab:s11}
\begin{tabular}{lccc}
\toprule
Correction & E & N & mean (px)\\
\midrule
raw (constant removed) & $1.390$ & $3.429$ & $1.205$\\
full-field destripe, $12^\circ$ (S2 value) & $0.251$ & $0.467$ & $0.179$\\
masked differential, $12^\circ$ & $0.241$ & $0.478$ & $0.180$\\
full-field destripe, $7.5^\circ$ (estimated) & $0.162$ & $0.311$ & $0.118$\\
\bottomrule
\end{tabular}
\end{table}

